\documentclass[lettersize,journal]{IEEEtran}
\usepackage{amsmath,amsfonts}
\usepackage{lmodern}
\usepackage{algorithm}
\usepackage{algpseudocode}
\usepackage{array}
\usepackage[caption=false,font=normalsize,labelfont=sf,textfont=sf]{subfig}
\usepackage{textcomp}
\usepackage{stfloats}
\usepackage{url}
\usepackage{verbatim}
\usepackage{graphicx}
\usepackage{cite}
\usepackage{pgfplots}

\usepackage{pifont}
\newcommand{\cmark}{\ding{51}} % Check mark
\newcommand{\xmark}{\ding{55}} % Cross mark
\begin{document}

\title{VENOM: Versatile Embodied Network for\\ Omni-bodied Motion tracking}

\author{Siddharth Padmanabhan$^{1}$, Kazuki Miyazawa$^{1}$, Takato Horii$^{1}$
        % <-this % stops a space

\thanks{$^{1}$Graduate School of Engineering Science,
        University of Osaka, 1-3 Machikaneyama-cho, Toyonaka-shi, Osaka 560-8531, Japan}%
}

% The paper headers
% \markboth{Journal of \LaTeX\ Class Files,~Vol.~14, No.~8, August~2021}%
% \markboth{IEEE Robotics and Automation Letters. June~2026}%
% {Shell \MakeLowercase{\textit{et al.}}: A Sample Article Using IEEEtran.cls for IEEE Journals}

% \IEEEpubid{0000--0000/00\$00.00~\copyright~2021 IEEE}
% Remember, if you use this you must call \IEEEpubidadjcol in the second
% column for its text to clear the IEEEpubid mark.

\maketitle

\begin{abstract}
Achieving expert-level expressive full-body motion tracking across multiple humanoids solely from demonstration data remains a challenging and relatively an underexplored problem in humanoid robot learning. Cross-embodiment motion tracking policies are mostly trained by decoupling the control problem into upper and lower body control. This work proposes VENOM, a cross-embodiment full-body motion tracking model for humanoids in simulation. VENOM is a GPT-based motion tracker trained on multiple humanoid data that can track the entire body without the requirement to split into upper and lower body control. We curate a multi-humanoid motion tracking dataset called the VENOM dataset that contains states, actions, and rewards and train VENOM and the baselines on this dataset. In this letter, we evaluate VENOM's performance against baselines and show that we can achieve a stable motion tracker across different humanoids more capable than an MLP trained on multiple humanoid data with supervised learning alone, and also show that despite lack of reward feedback, VENOM closely matches the tracking capability of experts that were trained using asymmetric-actor critic reinforcement learning.
\end{abstract}


\section{Introduction}
\IEEEPARstart{G}{eneralizable} cross-embodiment learning is an emerging paradigm in robot learning. By leveraging the capability of robot generalist policies to generalize across multiple modalities and adapt to unseen tasks \cite{bommasani2021opportunities} \cite{gpt3}, many researchers have explored crossing bodies and tasks via large-scale reinforcement learning. URMA\cite{bohlinger2024policyrunallendtoend} and URMA v2\cite{urmav2} have shown promise in developing robot-agnostic models and shows that it holds for scaling laws in increasing robot embodiments\cite{towards}. Models such as XHugWBC\cite{xhugwbc}, EAGLE\cite{eagle}, and LocoGPT\cite{locogpt}, all demonstrate the capability to control humanoids limited to locomotion-based task setting, deviating slightly from pure locomotion. However, these models have limited capability in expressing wide range behaviors and motions since motion retargeting across diverse humanoids still remains to be an open challenge.

\begin{figure*}
    \centering
    \includegraphics[width=1.0\linewidth, trim=0 120 0 120, clip]{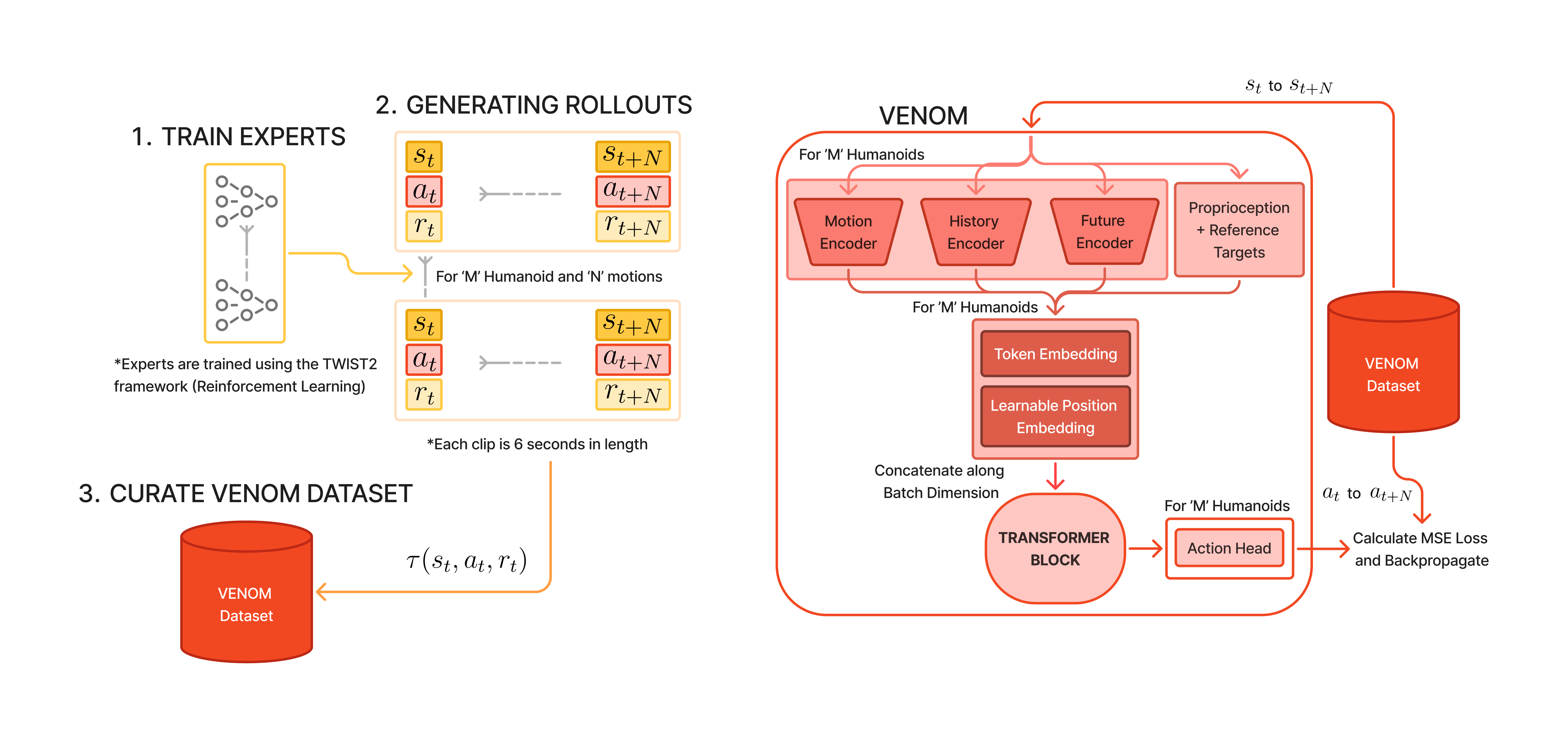}
    \caption{The flowchart on the left describes the pipeline to build the VENOM dataset. Experts are trained using TWIST2's training framework for several humanoids which is based on reinforcement learning. Considering there are $M$ humanoids and $N$ motions for each humanoid, $M*N$ successful trajectories are collected and split into clips of 6 seconds in length. The flowchart on the right describes the pipeline to train the VENOM model. Sequences of one second each are taken from the VENOM dataset for all the humanoids and fed to VENOM as input. The token embedding and position embedding are added for each humanoid input and concatenated along the batch dimension. This batch of input is passed to the decoder block and the outputs from the action heads are used to calculate the MSE losses. During inference, only one input module and action head is active, which is used to generate motion for one humanoid and task.}
    \label{fig:VENOM}
\end{figure*}

%%%% Proposal %%%%
We propose VENOM, a GPT-based multi-humanoid motion tracker, that can be switched between humanoids of different observation and action spaces for full-body motion tracking. We leverage the multi-tasking capability of GPT to train on a variety of behaviors and humanoids. The pipelines to curate the VENOM dataset and training VENOM are shown in Fig. \ref{fig:VENOM}. VENOM serves as the first multi-humanoid full-body motion tracking model. This serves as a starting point to generalize low-level full-body motion tracking for multiple humanoids.

%%%% Purpose of this Paper %%%%
We hypothesize that a GPT-based full-body multi-humanoid motion tracker, trained on diverse and noisy humanoid motion data using supervised learning, can demonstrate motion tracking across embodiments and motions far better than depriving data in terms of embodiments, noise in reference motion, and motion. We intend to demonstrate the successful form of foundation model training for full-body motion tracking in humanoids. 

%%%%%%%%%%%%%%%%%%%%%%%%%%%%%%%%%%%%%%%%%%%%%%%%%%%%%%%%%%%%
%%%% Contributions of this Paper %%%%
In this paper:
\begin{enumerate}
    \item we curate the VENOM dataset, a multi-humanoid dataset that consists of states, actions, and rewards collected in simulation via motion tracking,
    \item we demonstrate VENOM's ability to display full-body motion tracking between multiple humanoids and behaviors as a single GPT-based model,
    \item evaluate motion tracking capabilities between VENOM, MLP baselines, and Experts, and find that VENOM models outperform the MLP baselines trained on multiple humanoid data in joint motion tracking and root stability across different behavior categories, and
    \item we quantify how close supervised learning can get to RL experts.
\end{enumerate}

\section{Related Literature}
\begin{table*}
    \centering
    \caption{Comparison of state-of-the-art models (RL stands for Reinforcement Learning)}
    \label{tab:crossmodels}
    \begin{tabular}{c|c|c|c|c|c}
        \hline
        Model    & Multi-      & Unified  & Expressive    & Core Controller & Dataset \\
                 & Humanoid    &   WBC   & Behaviors     &                 &         \\
        \hline
        MoCapAct \cite{mocapact} & \xmark      & \cmark       & \cmark        & Transformer & MoCapAct Dataset                \\
        URMA     \cite{bohlinger2024policyrunallendtoend} & \cmark      & \cmark       & \xmark        & MLP         & RL                              \\
        TWIST2   \cite{twist2} & \xmark      & \cmark       & \cmark        & MLP         & RL with AMASS, OMOMO            \\
        SONIC    \cite{sonic}  & \xmark      & \cmark       & \cmark        & MLP         & RL with Humanoid Motion Dataset \\
        HMG      \cite{padmanabhan} & \xmark      & \cmark       & \cmark        & Transformer & MoCapAct Dataset                \\
        LocoGPT  \cite{locogpt} & \cmark      & \cmark       & \xmark        & Transformer & LocoMuJoCo                      \\
        EAGLE    \cite{eagle} & \cmark      & \xmark       & \xmark        & MLP         & RL                              \\
        XHugWBC  \cite{xhugwbc} & \cmark      & \xmark       & \xmark        & Transformer & RL                              \\
        VENOM (ours) & \cmark  & \cmark       & \cmark        & Transformer & VENOM Dataset                   \\
        \hline
    \end{tabular}
\end{table*}

\subsection{Robot Generalists Policies}
Foundation models and robot generalist policies have been trending for a while in robotics because of their high versatility. Meta Motivo\cite{tirinzoni2024metamotivo} and its successor BFM-Zero\cite{bfmzero} are good examples of a behavioral foundation model trained on single humanoids (including Unitree's G1) using unsupervised reinforcement learning that can be prompted to solve motion tracking, goal reaching, and reward optimization. Octo is another example of a robot generalist policy that is trained on the Open X-Embodiment dataset, that can be fine-tuned to tasks that have different observation and action spaces in a flexible manner for robot manipulation \cite{octo}. Although a lot of research is promoted on Unitree's G1, for other humanoids, unlike datasets like Open X-Embodiment dataset \cite{rtx}, there is a significant lack of large-scale low-level control data available to train versatile foundation models. In this work, we curate our own large-scale teleoperation dataset and train our motion tracker on it.

\subsection{Cross-Embodiment Models}
Lately there has been active research in cross-embodiment where controllers can generalize across different embodiments that vary in observation and action spaces. Perpetual Humanoid Control\cite{perpetual} is one of the earliest works in this domain, where a progressive neural network model is trained on multiple behaviors on several SMPL bodies with the same observation and action spaces. Body Transformer\cite{sferrazza2024bodytransformerleveragingrobot} uses graphical representations to represent robots as graphs of sensors and actuators. In \cite{universalhumanoid} researchers propose a framework that can automatically discover valid humanoid character based on target human sequences. CrossFormer \cite{crossformer} is a versatile transformer-based model that can consume data from any robot morphology, alleviating the need of manually aligning observation and action spaces. URMA\cite{bohlinger2024policyrunallendtoend} and its successor URMA v2\cite{urmav2} are embedding algorithms proposed to alleviate manual aligning of observation and action spaces for legged locomotion. Some of the recent works like EAGLE\cite{eagle} and XHugWBC\cite{xhugwbc} propose controllers that track body velocities and height to perform a variety of locomotion-based behaviors. EAGLE implements policy distillation from different experts and trains the policy using reinforcement learning, whereas XHugWBC scales in the embodiment direction directly from scratch in reinforcement learning environment.  However, there are no cross-embodiment controllers that have been directly trained on large-scale human motion capture datasets like AMASS\cite{amass}, BABEL\cite{babel}, and OMOMO\cite{omomo}. In this work, we use the AMASS dataset to generate retargeted motions for five humanoids and use this as our source of large-scale motion-tracking data.

\subsection{Humanoid Teleoperation}
One of the most common applications of humanoid generalist policies is in teleoperation. There has been a surge in teleoperation frameworks in recent years. The earliest works demonstrated teleoperation on Unitree H1 \cite{he2024hover, fu2024humanplus, he2024learning} by leveraging large-scale imitation learning with student-teacher training and decoupling upper and lower body control. This progressed into full-body motion tracking on Unitree G1 \cite{beyondmimic, clone, twist, twist2, sonic}. We leverage the training pipeline used in TWIST2\cite{twist2} to curate the VENOM dataset and we use the IsaacGym pipeline to test and evaluate VENOM, its counterparts and baselines. 

\subsection{Datasets}
Popular motion capture datasets for human motion that are available to the public like AMASS, Human3.6M, OMOMO, BABEL \cite{h36, babel, omomo, amass} are extensively used in imitation learning for humanoids. There are limited number of datasets that provide low-level control data for humanoids, such as MoCapAct\cite{mocapact} and LocoMuJoCo datasets\cite{locomujoco}. In this work, we curate the VENOM dataset by collecting successful motion trajectories from expert motion trackers trained using TWIST2's training pipeline\cite{twist2}. Unlike prior approaches that train dedicated controllers for each motion and humanoid\cite{mocapact, locomujoco}, the VENOM dataset was generated using small number of embodiment-specific experts rather than requiring thousands of humanoid-motion-specific experts. This substantially reduces the computational burden of dataset generation while enabling collection of diverse full-body motion tracking data across multiple humanoids and behaviors. A summary of related work and comparison of VENOM with other models is given in the form of a table (Table \ref{tab:crossmodels}). Expressive behaviors in this table are behaviors that are not restricted to combination of basic locomotion, squatting, and manipulation, but extends to other dynamic motions that include motions with flight phases such as running, jumping, and dancing.

\section{Method}
We introduce VENOM, a versatile embodied network for omni-bodied motion tracking. This problem is framed as a full-body motion tracking task for multiple humanoids. VENOM is built on the training pipeline for training an autoregressive GPT model used in MoCapAct \cite{mocapact}, training pipeline for training motion tracking policy used in TWIST2 \cite{twist2}, and the model architecture used in LocoGPT \cite{locogpt}, which are described in detail in the following subsections. The pipeline for building VENOM dataset and the pipeline for training VENOM is shown in Fig. \ref{fig:VENOM}.

\subsection{Dataset}
The pipeline for curating the VENOM dataset is built on the same approach used to build the  MoCapAct dataset \cite{mocapact}. We curated the VENOM dataset from five different humanoids: Unitree G1, Unitree H1, Unitree H1 v2, Booster T1 with 29 degrees of freedom, and Fourier N1. Expert motion trackers were trained on the entirety of the AMASS dataset for each humanoid robot using the TWIST2's training pipeline\cite{twist2}. Next, these experts were used to generate one clean and eight noisy rollouts that consisted of observations, actions, and rewards for each motion in the AMASS dataset. Noise was applied to reference motion in observation; $0.01$ to $0.05$ meter noise range for root position, $0.1$ to $0.2$ radian noise range for root orientation, $0.05$ to $0.1$ meter per second noise range for root velocity and $0.05$ to $0.1$ radian noise range for joint position. All of these rollouts were split into clips of 6 seconds each. The dataset contains rollouts of five different humanoid expert motion trackers which amounts to roughly $7705$ hours of data. The details of the dataset are given in Table \ref{tab:venomdataset}. For Unitree H1 v2, only 50 hours of training data and 16 hours of validation data were retained. Expert for Unitree H1 v2 generated trajectories that had lower success rates particularly when noise was introduced in reference motion. Hence to avoid using unreliable demonstrations we restricted the dataset for this platform. We later see in the results that despite this we observe no meaningful difference in the downstream performance.

\subsection{Model}
We build on the architectural design of LocoGPT \cite{locogpt} as the motion tracker, consisting of multiple input embeddings and action heads that correspond to each humanoid with a transformer decoder architecture sitting in the center. The core controller of VENOM has eight transformer layers and eight attention heads in each layer with an embedding dimension of 768. The primary distinction between LocoGPT and VENOM lies in their application, where LocoGPT was formulated for autoregressive motion generation, whereas VENOM is formulated as a full-body motion tracker, that includes motion, history, and future encoders inspired by TWIST2 to support this objective.

For comparative evaluation, we use MLP architectures implemented in TWIST2 as the baseline model. The network consists of six layers: the input layer corresponding to the padded observation, following three layers of size 256, a 256-dimensional LayerNorm, and an output layer corresponding to the action dimension.

In addition, there are motion encoders, history encoders, and future encoders that are used to encode the reference motions before sending them to the main network in VENOM and TWIST2. The motion, history, and future encoders are used to encode the one step reference motion, reference motion history of ten steps, and one step future reference motion into compact latent representations.

For the motion encoder, the input is projected into a 60-dimensional linear layer and the resulting features are flattened and passed through the final layer to produce latent motion representations of latent dimension 128.

For the history encoder, similar to the motion encoder, the input is projected into a 60-dimensional linear layer and then processed using two 1-D convolutions with kernel sizes and strides being (4,2) and (2,1), respectively. The resulting features are flattened and passed through the final layer to produce latent history representations of latent dimension 128.

For the future encoder, unlike the previous encoders, no convolutions are used. The input is passed through a 3-layered MLP with hidden dimension 256, 128. Dropout regularization is applied between hidden layers to improve generalization.
The activation function used in the baselines and encoders are all SiLU. The summary of hyperparameters are given in Table \ref{tab:hyperparameters}.

\begin{table}
    \centering
    \caption{Model Architecture and Hyperparameters}
    \label{tab:hyperparameters}
    \begin{tabular}{c|c}
        \hline
        \multicolumn{2}{c}{VENOM Hyperparameters} \\
        \hline
        Transformer Layers & 8 \\
        Attention Heads    & 8 \\
        Embedding Dim & 768 \\
        Block Size & 50 \\
        Learning Rate & 0.0001 \\
        Batch Size (per robot ) & 8 \\
        Optimizer & AdamW \\
    \end{tabular}
    \begin{tabular}{c|c}
        \hline
        \multicolumn{2}{c}{TWIST2 Hyperparameters} \\
        \hline
        Layers & 6 \\
        Hidden Layer Dim & 256 \\
        No. of Layer Norm & 1 \\
        Layer Norm Dim & 256 \\
        Activation & SiLU \\
        Learning Rate & 0.0001 \\
        Batch Size (per robot) & 16 \\
        Optimizer & AdamW \\
    \end{tabular}
    \begin{tabular}{c|c}
        \hline
        \multicolumn{2}{c}{Motion Encoder} \\
        \hline
        Motion Encoder & 1 Linear Layer \\
        Motion Observations & 1 step \\
        Latent Dim & 128 \\
    \end{tabular}
    \begin{tabular}{c|c}
        \hline
        \multicolumn{2}{c}{History Encoder} \\
        \hline
        History Encoder & 1-D Convolution \\
        Convolution Layers & 2 \\
        Kernel Size & (4,2), (2,1) \\
        History Observations & 10 steps \\
        Latent Dim & 128 \\
    \end{tabular}
    \begin{tabular}{c|c}
        \hline
        \multicolumn{2}{c}{Future Encoder} \\
        \hline
        Future Encoder & MLP \\
        MLP Layers & 3 \\
        MLP Hidden Dim & 256 \\
        Future Observations & 1 step \\
        Latent Dim & 128 \\
    \end{tabular}
\end{table}

\subsection{Training}
Similar to LocoGPT, for respective input modules, token embedding layers are represented by feedforward layers. These layers process each observation data (batch size $B$ and observation size $O$) into token embeddings of embedding dimension $D$. These are added to the learned position embeddings. Ultimately, the observation data transform from dimensions $[B, T, O]$ to $[B, T, D]$ by the time they reach the transformer block. 

Before entering the transformer block, the output of each input module is concatenated along the batch dimension $B$ and sent as one batch to the core transformer block. The output of the transformer block is distributed and sent to their respective action heads where each is a feedforward linear layer that outputs actions for respective humanoids. However, during inference, the input module and action head that correspond to the humanoid of interest are used, and the rest are ignored.

VENOM, individual motion tracking experts, and the TWIST2 baselines and encoders were trained for 2M steps and the learning rate used here is $1*10^{-4}$. Context lengths of 50 steps (1 second) are used as motion prompts. Each motion prompt contains 50 step sequence of humanoid state-observation. The input to the policy is updated using a sliding-window mechanism by popping the first state observation of the motion prompt and appending the new state observation at the end. The action heads output sequence of actions from which the actions of the last step are taken and sent to the robot for control.

Observations are normalized using running estimates of feature-wise mean and standard deviation. The statistics such as sample count, feature-wise sum, and squared sum, are updated online while training to compute running estimates of mean and variance, and normalized values are clipped to improve numerical stability. The actions are clipped between -10 to 10 in magnitude, as done in TWIST2's implementation.

Algorithm \ref{alg:venomtrain} and Algorithm \ref{alg:venominf} describe training and inference procedures, respectively.

The weights of VENOM and baselines were updated by minimizing MSE loss. Observations consists of current joint targets, joint positions, joint velocities, ten steps of proprioception history, and one step of future joint targets. The actions are sent to the humanoids as PD targets. The simulator used in this work to test VENOM and generate the dataset is IsaacGym \cite{isaacgym}.

\begin{table}
    \centering
    \caption{VENOM Dataset}
    \label{tab:venomdataset}
    \begin{tabular}{c|c|c|c}
        \hline
        Robot & Training & Validation & State \\
         & Dataset & Dataset & (Action) \\
         & Size (hours) & Size (hours) & Size \\
        \hline
        Unitree G1        & 1363 & 343 & 1432(29)\\
        Unitree H1        & 1590 & 400 & 982(19)\\
        Unitree H1 v2     & 50   & 16  & 1342(27)\\
        Fourier N1        & 1546 & 387 & 1432(23) \\
        Booster T1 29 DoF & 1607 & 403 & 1342(27) \\
        \hline
        Total & 6156 & 1549
    \end{tabular}
\end{table}

\begin{figure}
    \centering
    \begin{tikzpicture}        
        % \node[rotate=90, scale=0.6] at (-3.5,-3.0) {\fontfamily{lmss}\selectfont Unitree G1};
        \node at (-10.0,-4.0) {\includegraphics[trim={0.6cm 2cm 2cm 2cm}, clip, scale=0.06]{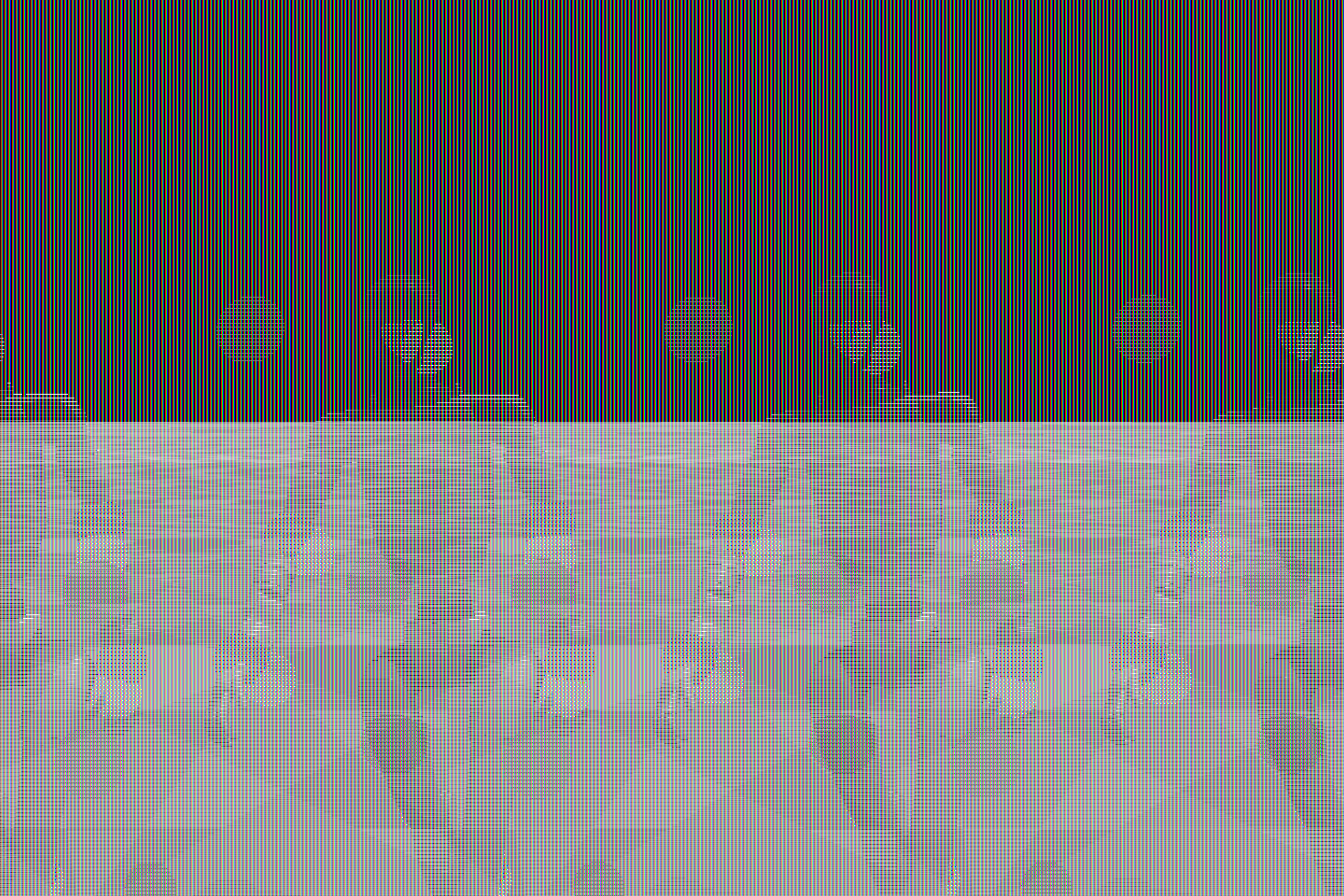}};
        \hspace{0pt}%
        % \node at (0.5,-3.0) {\includegraphics[trim={0.6cm 2cm 2cm 2cm}, clip, scale=0.05]{figures/frames/G1/env0_frame_000140.pdf}};
        % \hspace{0pt}%
        % \node at (3.0,-3.0) {\includegraphics[trim={0.6cm 2cm 2cm 2cm}, clip, scale=0.05]{figures/frames/G1/env0_frame_000200.pdf}};

        % \node[rotate=90, scale=0.6] at (-3.5,-4.6) {\fontfamily{lmss}\selectfont Unitree H1};
        \node at (-7.0,-4.0) {\includegraphics[trim={0.6cm 2cm 2cm 2cm}, clip, scale=0.06]{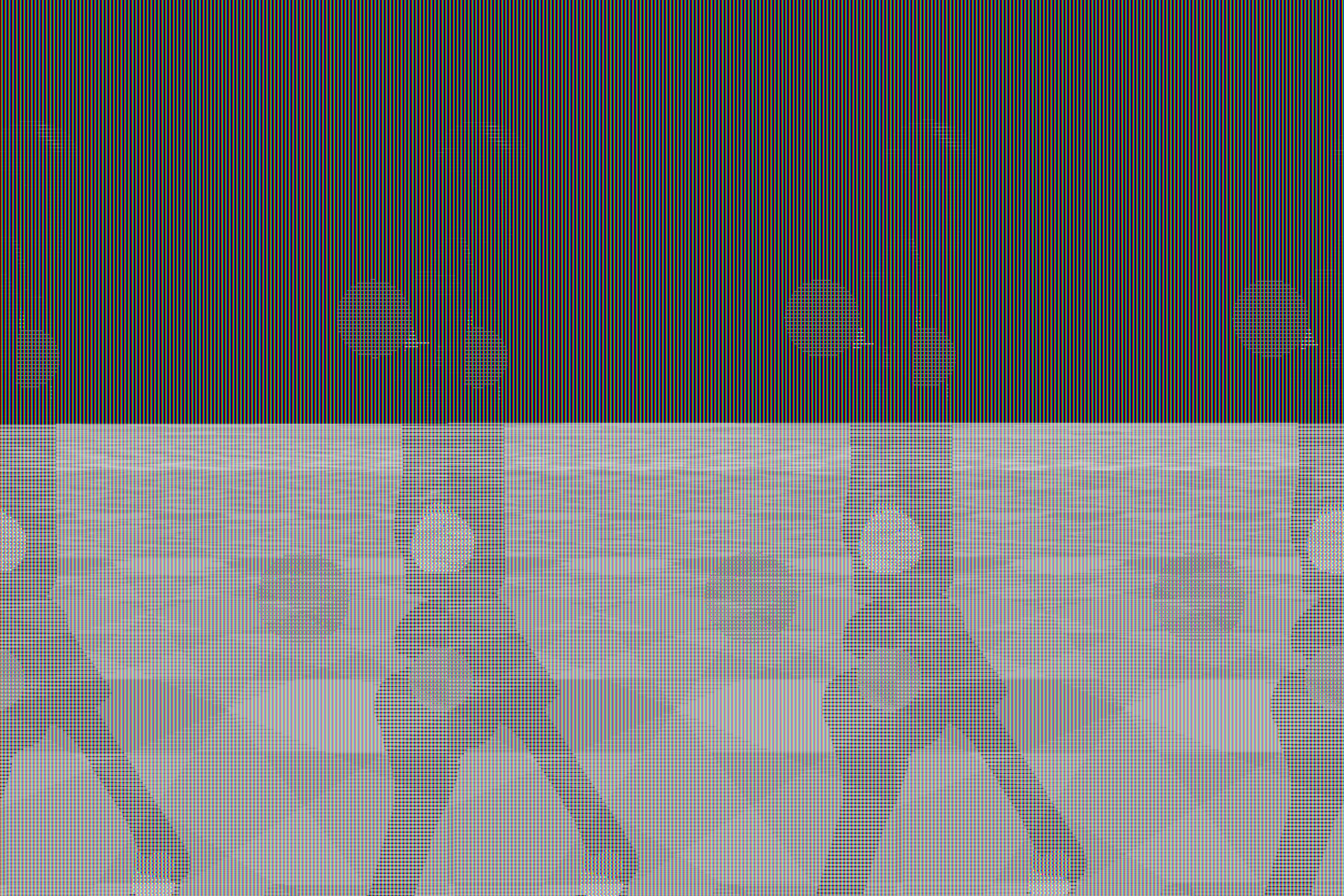}};
        \hspace{0pt}%
        % \node at (0.5,-4.6) {\includegraphics[trim={0.6cm 2cm 2cm 2cm}, clip, scale=0.05]{figures/frames/H1/env0_frame_000140.pdf}};
        % \hspace{0pt}%
        % \node at (3.0,-4.6) {\includegraphics[trim={0.6cm 2cm 2cm 2cm}, clip, scale=0.05]{figures/frames/H1/env0_frame_000200.pdf}};

        % \node[rotate=90, scale=0.6] at (-3.5,-6.2) {\fontfamily{lmss}\selectfont Unitree H1 v2};
        \node at (-4.0,-4.0) {\includegraphics[trim={0.6cm 2cm 2cm 2cm}, clip, scale=0.06]{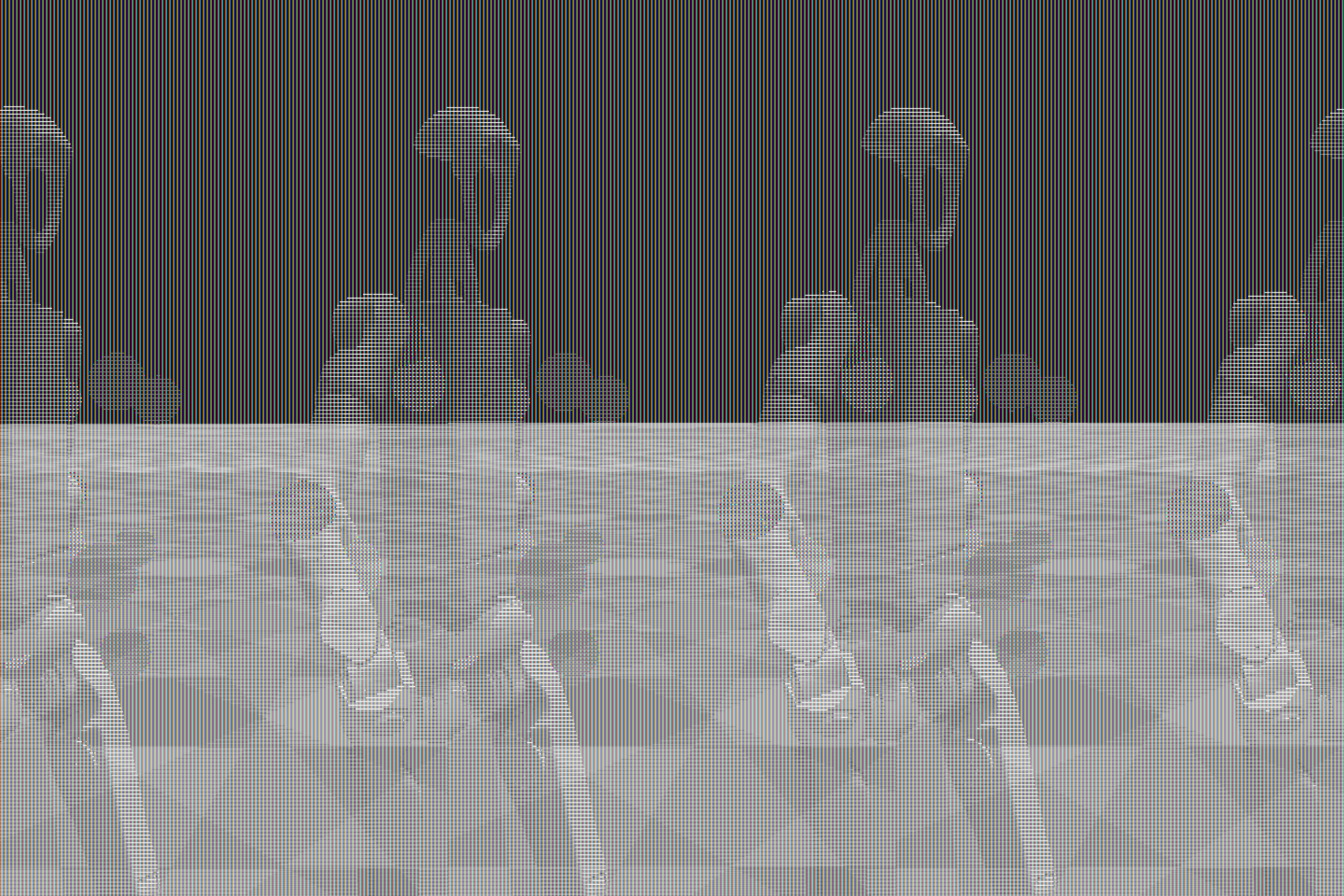}};
        \hspace{0pt}%
        % \node at (0.5,-6.2) {\includegraphics[trim={0.6cm 2cm 2cm 2cm}, clip, scale=0.05]{figures/frames/H12/env0_frame_000140.pdf}};
        % \hspace{0pt}%
        % \node at (3.0,-6.2) {\includegraphics[trim={0.6cm 2cm 2cm 2cm}, clip, scale=0.05]{figures/frames/H12/env0_frame_000200.pdf}};

        % \node[rotate=90, scale=0.6] at (-3.5,-7.8) {\fontfamily{lmss}\selectfont Fourier N1};
        \node at (-8.5,-6.0) {\includegraphics[trim={0.6cm 2cm 2cm 2cm}, clip, scale=0.06]{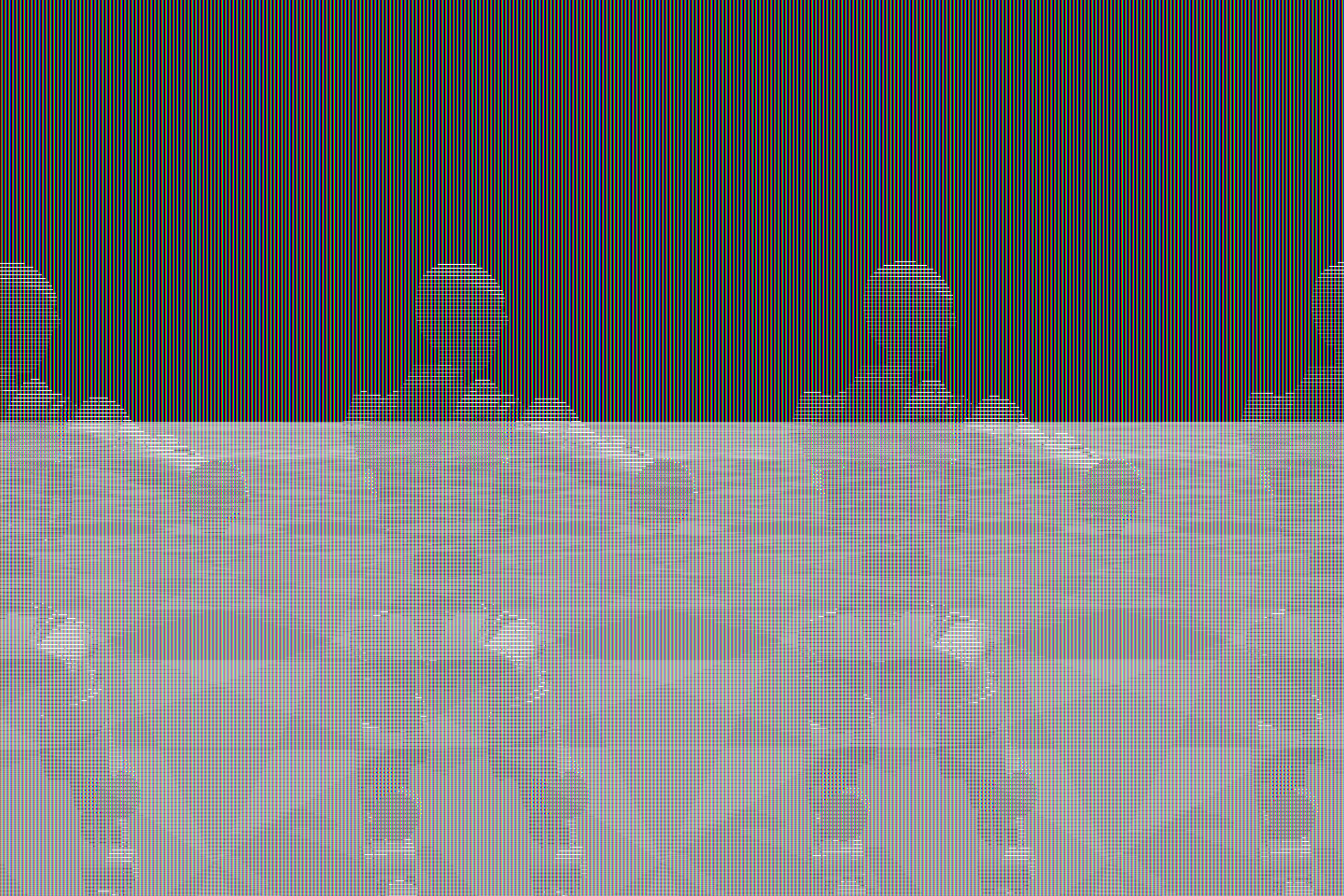}};
        \hspace{0pt}%
        % \node at (0.5,-7.8) {\includegraphics[trim={0.6cm 2cm 2cm 2cm}, clip, scale=0.05]{figures/frames/N1/env0_frame_000140.pdf}};
        % \hspace{0pt}%
        % \node at (3.0,-7.8) {\includegraphics[trim={0.6cm 2cm 2cm 2cm}, clip, scale=0.05]{figures/frames/N1/env0_frame_000200.pdf}};

        % \node[rotate=90, scale=0.6] at (-3.8,-9.4) {\fontfamily{lmss}\selectfont Booster T1};
        % \node[rotate=90, scale=0.6] at (-3.5,-9.4) {\fontfamily{lmss}\selectfont 29 DoF};
        \node at (-5.5,-6.0) {\includegraphics[trim={0.6cm 2cm 2cm 2cm}, clip, scale=0.06]{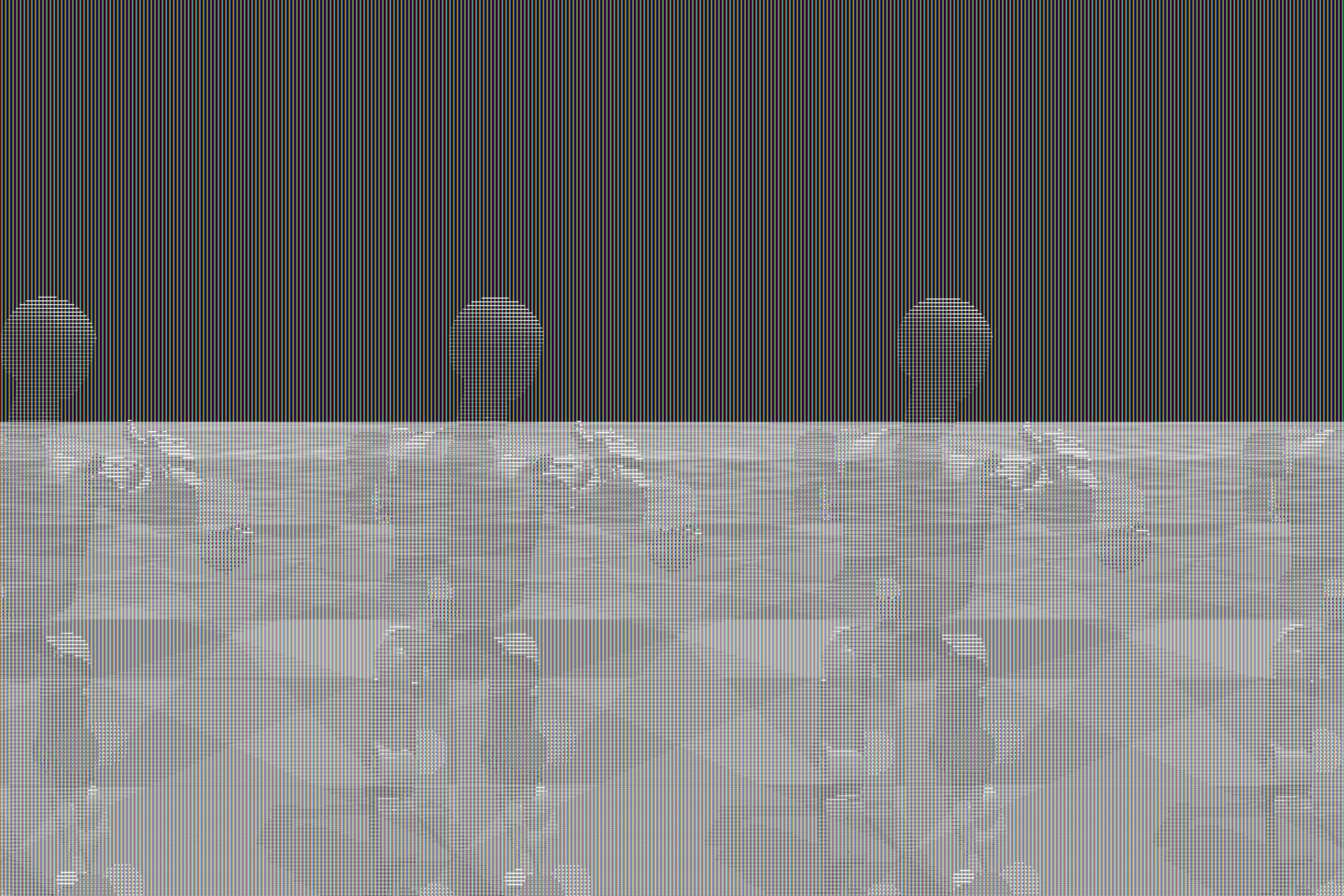}};
        \hspace{0pt}%
        % \node at (0.5,-9.4) {\includegraphics[trim={0.6cm 2cm 2cm 2cm}, clip, scale=0.05]{figures/frames/T129/env0_frame_000140.pdf}};
        % \hspace{0pt}%
        % \node at (3.0,-9.4) {\includegraphics[trim={0.6cm 2cm 2cm 2cm}, clip, scale=0.05]{figures/frames/T129/env0_frame_000200.pdf}};
    \end{tikzpicture}

    \caption{The above frames are generated by using VENOM-noisy for the five different robots for different motions using the TWIST2 inference pipeline. The red markers are the humanoids current joint positions, the blue markers are the local reference positions, and the green markers are the global reference positions. Top (from left to right): Unitree G1, Unitree H1, Unitree H1 v2, Bottom (left to right): Fourier N1 and Booster T1 29DoF}
    \label{fig:screenshots}
\end{figure}

\begin{algorithm}
    \caption{VENOM Omni-Bodied Motion Tracking Training}
    \label{alg:venomtrain}
    \begin{algorithmic}[1]
        \Require Context length $T = 50$, Number of Humanoids $N_{H} = 5$, Body-Specific Encoded State Trajectories $\{\tau_b\}_{b=1}^{N_{H}}$ where $\tau = \{s_t\}_{t=0}^T$, Model Parameters $\theta$, VENOM Motion Tracking Policy $f_\theta$, Token Embedding Layers $\{W_{\theta, b}\}_{b=1}^{N_{H}}$, Learnable Positional Embedding $\{P_{\theta, b}\}_{b=1}^{N_{H}}$, Output Modules $\{g_{\theta, b}\}_{b=1}^{N_{H}}$, MSE Loss Function $\mathcal{L}$, Loss Sum $\ell_{sum} = 0$, Average Loss $\ell_{avg}$

        \Ensure Trajectories of Actions $\{{A}_b\}_{b=1}^{N_{H}}$, where $A = \{a_t\}_{t=0}^{T}$

        \For{each iteration until convergence}
            \For {$b = 1$ to ${N_{H}}$}
            \State Construct input embedding $C_b \gets W_{\theta, b}(\tau_b) + P_{\theta, b}$
            \EndFor

            \State Concatenate contexts along batch dimension:
            \Statex \hspace{\algorithmicindent} $C \gets \text{Concat}(\{C_b\}_{b=1}^{N_{H}})$

            \State Pass the concatenated input to the transformer block:
            \Statex \hspace{\algorithmicindent} $Z \gets f_\theta(C)$

            \For {$b = 1$ to ${N_{H}}$}
            \State Segregated body-specific output along the batch dimension $Z_b$ from $Z$
            \State Compute predicted trajectory of actions $\hat{A}_b \gets g_b(Z_b)$
            \State Compute loss sum $\ell_{sum} \gets \ell_{sum} + \mathcal{L}(\hat{A}_b, A_b)$
            \EndFor
            \State Compute loss average $\ell_{avg} \gets \ell_{sum}/N_{H}$
            \State Update parameters $\theta \gets \theta - \eta \nabla_\theta \ell_{avg}$
        \EndFor
    \end{algorithmic}
\end{algorithm}

\begin{algorithm}
    \caption{VENOM Omni-Bodied Motion Tracking Inference}
    \label{alg:venominf}
    \begin{algorithmic}[1]
        \Require Context length $T = 50$, Encoded State Trajectory $\tau$ where $\tau = \{s_t\}_{t=0}^T$, Model Parameters $\theta$, VENOM Motion Tracking Policy $f_\theta$,  Humanoid Index $N$, Token Embedding Layers $\{W_{\theta, N}\}$, Learnable Positional Embedding $\{P_{\theta, N}\}$, Output Modules $\{g_{\theta, N}\}$

        \Ensure Trajectories of Actions $A$, where $A = \{a_t\}_{t=0}^{T}$
        \While {Episode Not Terminated}
            \State Construct input embedding $C \gets W_{\theta, N}(\tau) + P_{\theta, N}$
            \State Pass the input to the transformer block:
            \Statex \hspace{\algorithmicindent} $Z \gets f_\theta(C)$
            \State Compute predicted trajectory of actions $\hat{A} \gets g_N(Z)$
            \State Send actions $A_N$ to Humanoid N
            \State Update trajectory:
        $\tau \gets \text{PopFront}(\tau) \cup \{s_{(t+1)}\}$
        \EndWhile
    \end{algorithmic}
\end{algorithm}

% \begin{figure}
%     \centering
%     \includegraphics[width=1.0\linewidth]{figures/g1_all_tracking_error.pdf}
%     \caption{This plot shows the tracking error in Unitree G1 between joint positions and desired joint positions for the VENOM and TWIST2 variants. We can see that TWIST2's architecture cannot handle multi-robot data as well as the VENOM variants.}
%     \label{fig:trackingerror}
% \end{figure}

\begin{figure*}
    \centering
    \begin{tikzpicture}        
        \node[rotate=90, scale=0.6] at (-4.0,-3.0) {\fontfamily{lmss}\selectfont VENOM-noisy};
        \node at (-2.0,-3.0) {\includegraphics[trim={0.6cm 2cm 2.6cm 2cm}, clip, scale=0.07]{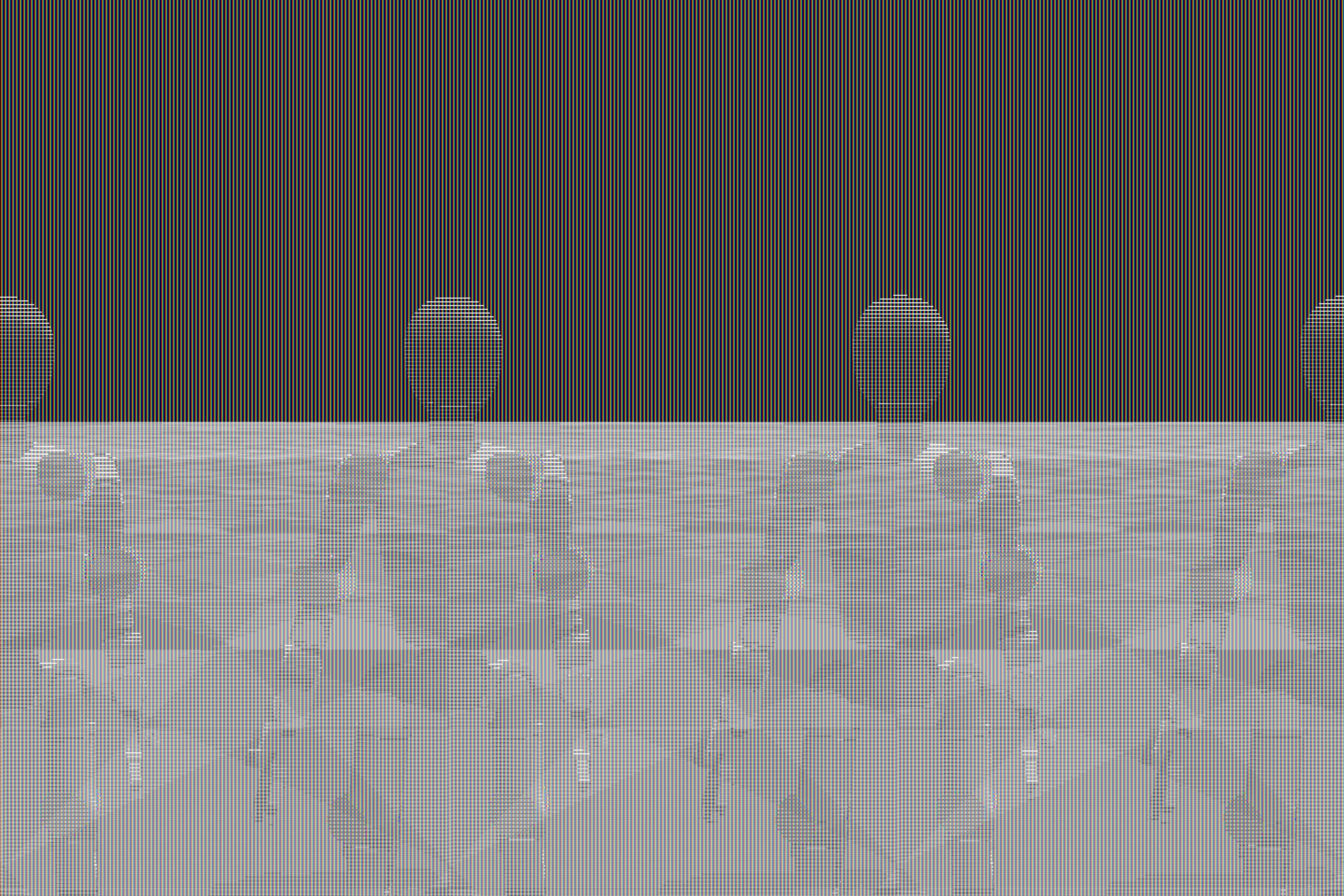}};
        \hspace{0pt}%
        \node at (1.5,-3.0) {\includegraphics[trim={0.6cm 2cm 2cm 2cm}, clip, scale=0.07]{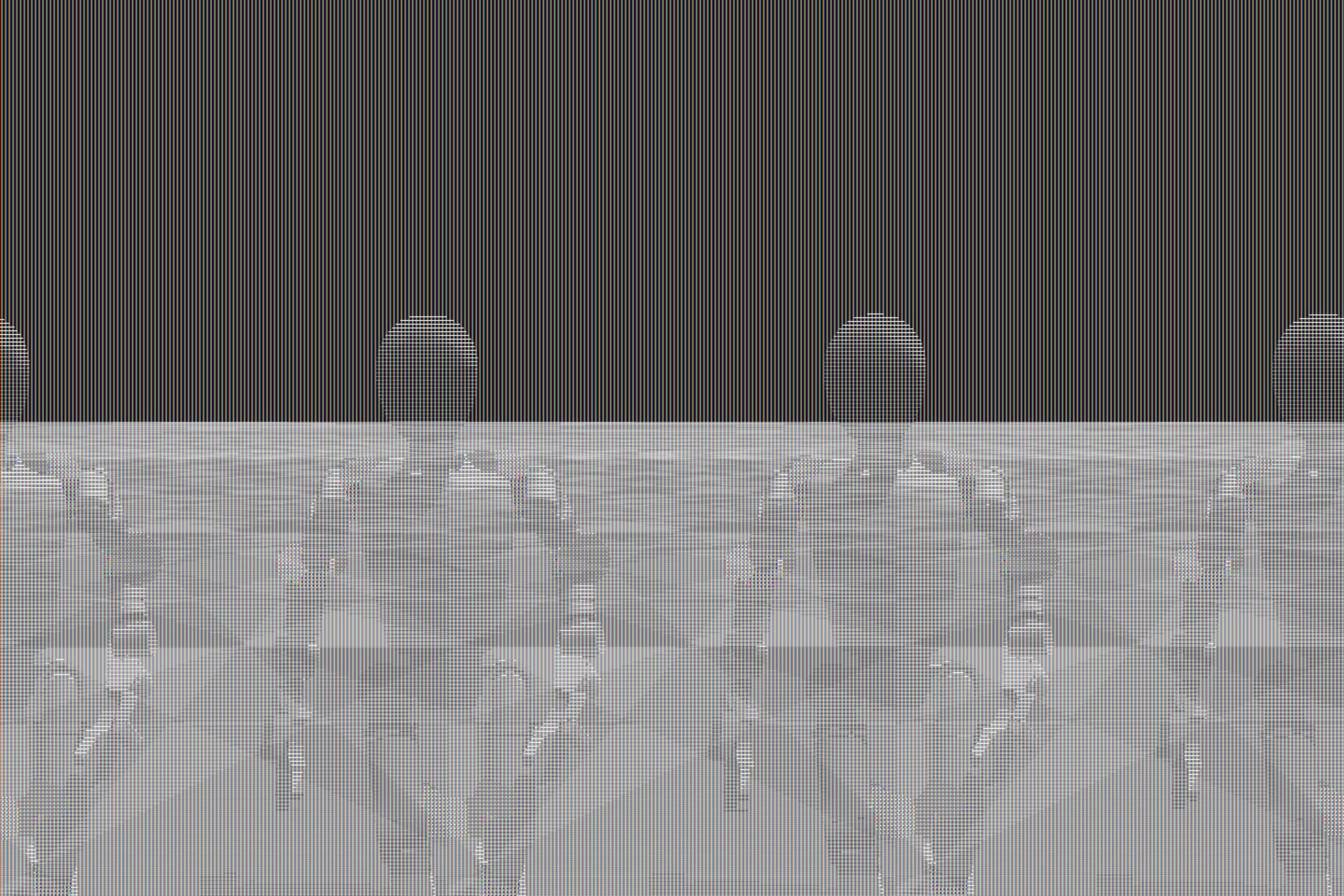}};
        \hspace{0pt}%
        \node at (5.0,-3.0) {\includegraphics[trim={0.6cm 2cm 2cm 2cm}, clip, scale=0.07]{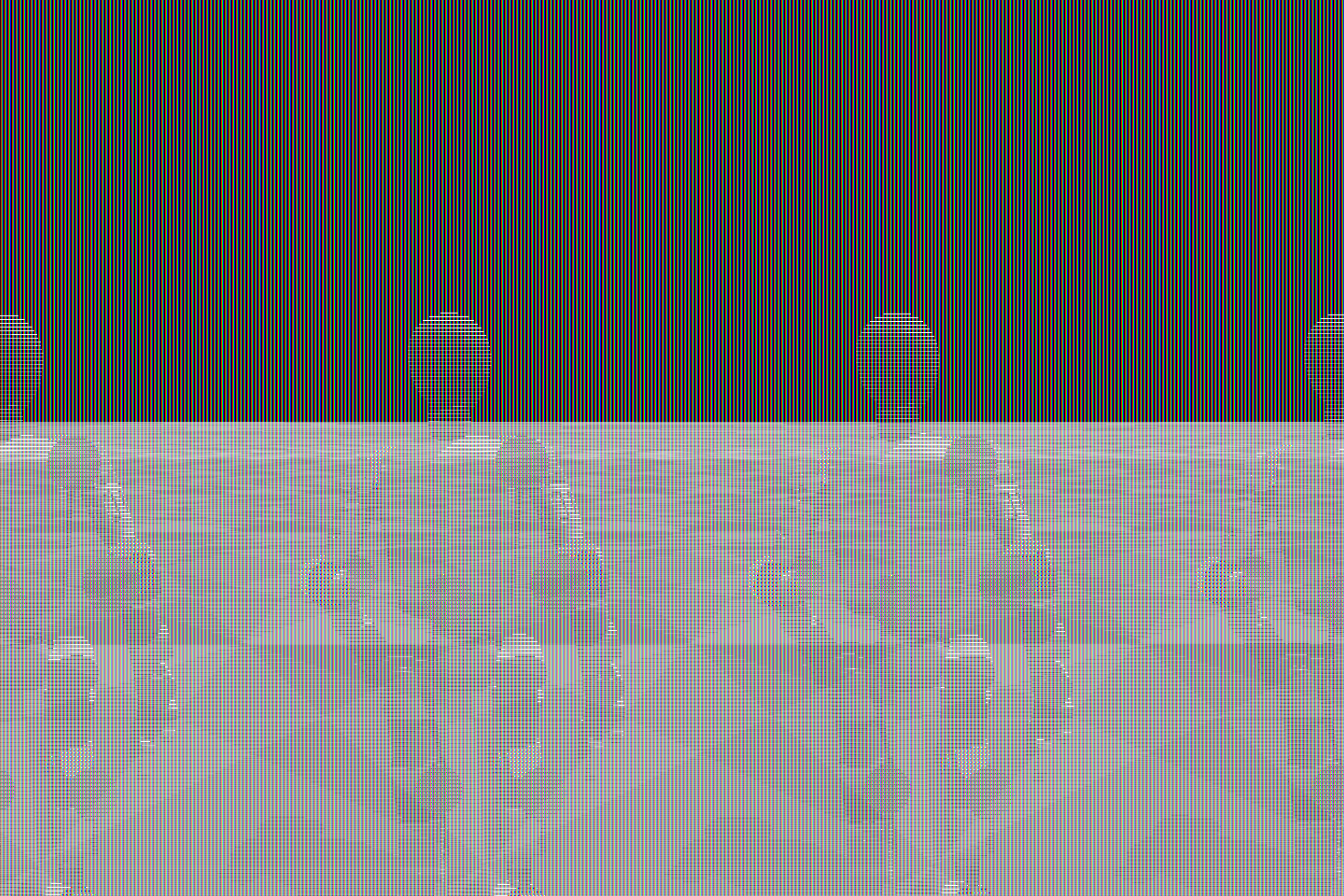}};
        \hspace{0pt}%
        \node at (8.5,-3.0) {\includegraphics[trim={0.6cm 2cm 2cm 2cm}, clip, scale=0.07]{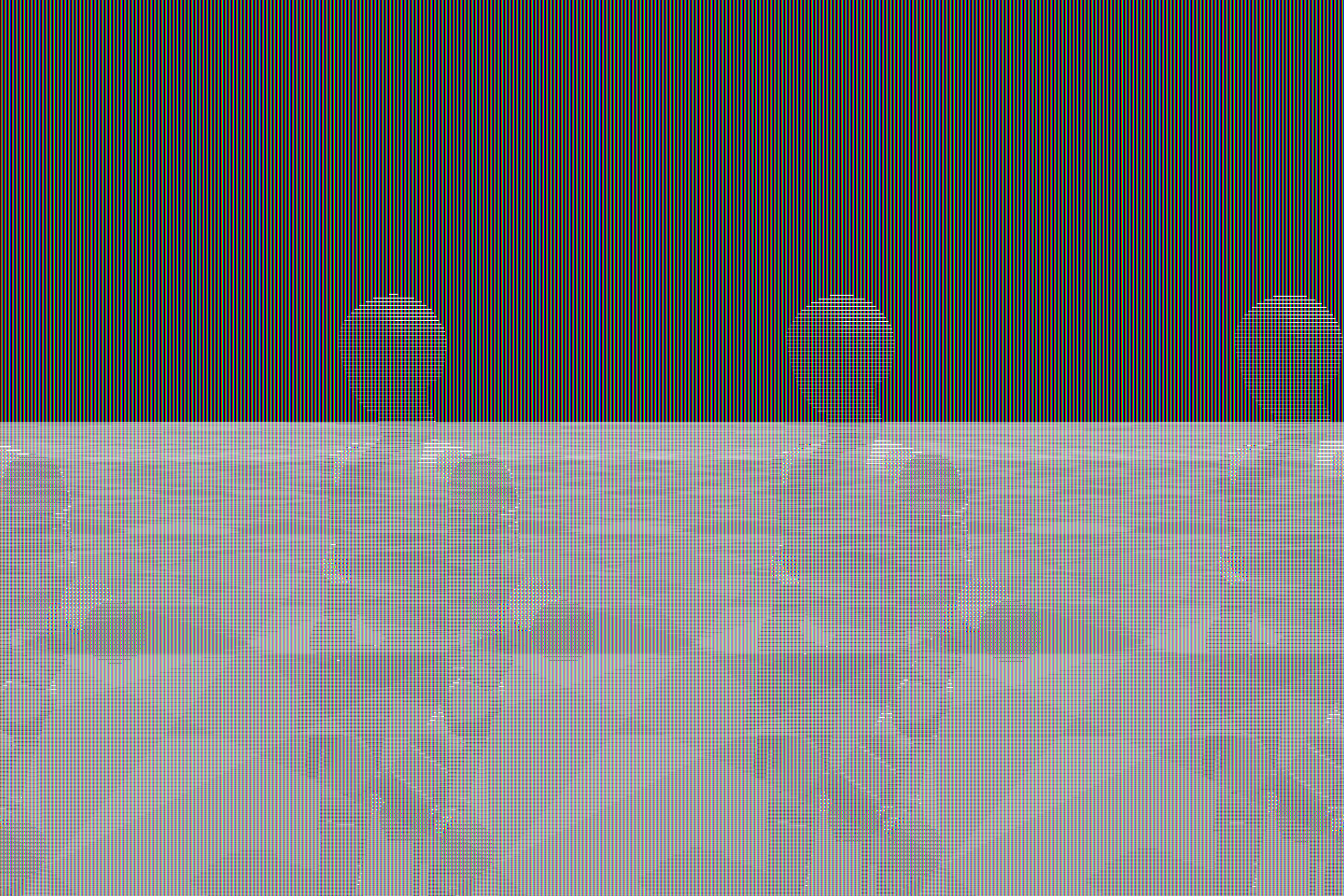}};
        \hspace{0pt}%
        \node at (12.0,-3.0) {\includegraphics[trim={0.6cm 2cm 2cm 2cm}, clip, scale=0.07]{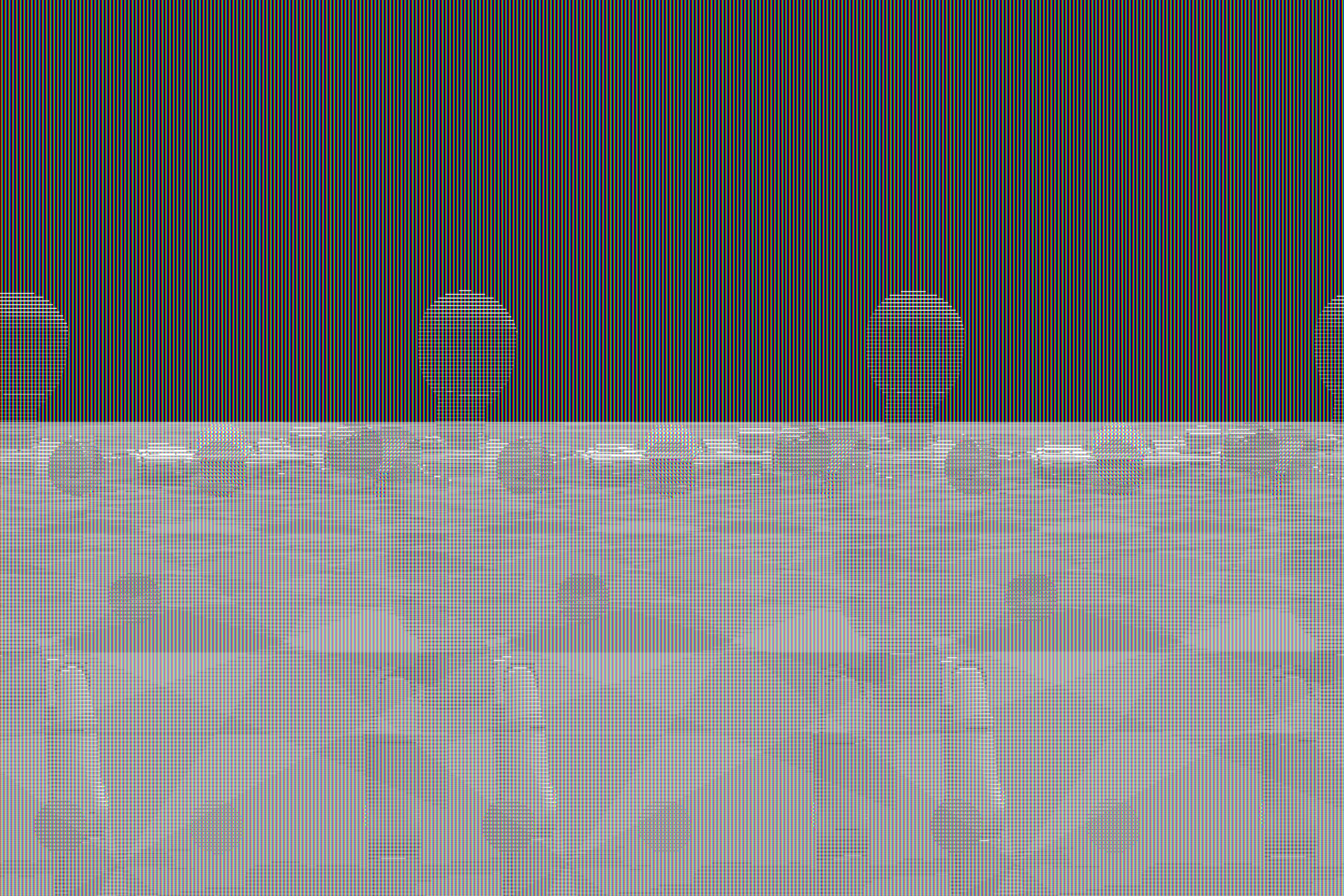}};

        \node[rotate=90, scale=0.6] at (-4.0,-5.2) {\fontfamily{lmss}\selectfont VENOM-clean};
        \node at (-2.0,-5.2) {\includegraphics[trim={0.6cm 2cm 2.6cm 2cm}, clip, scale=0.07]{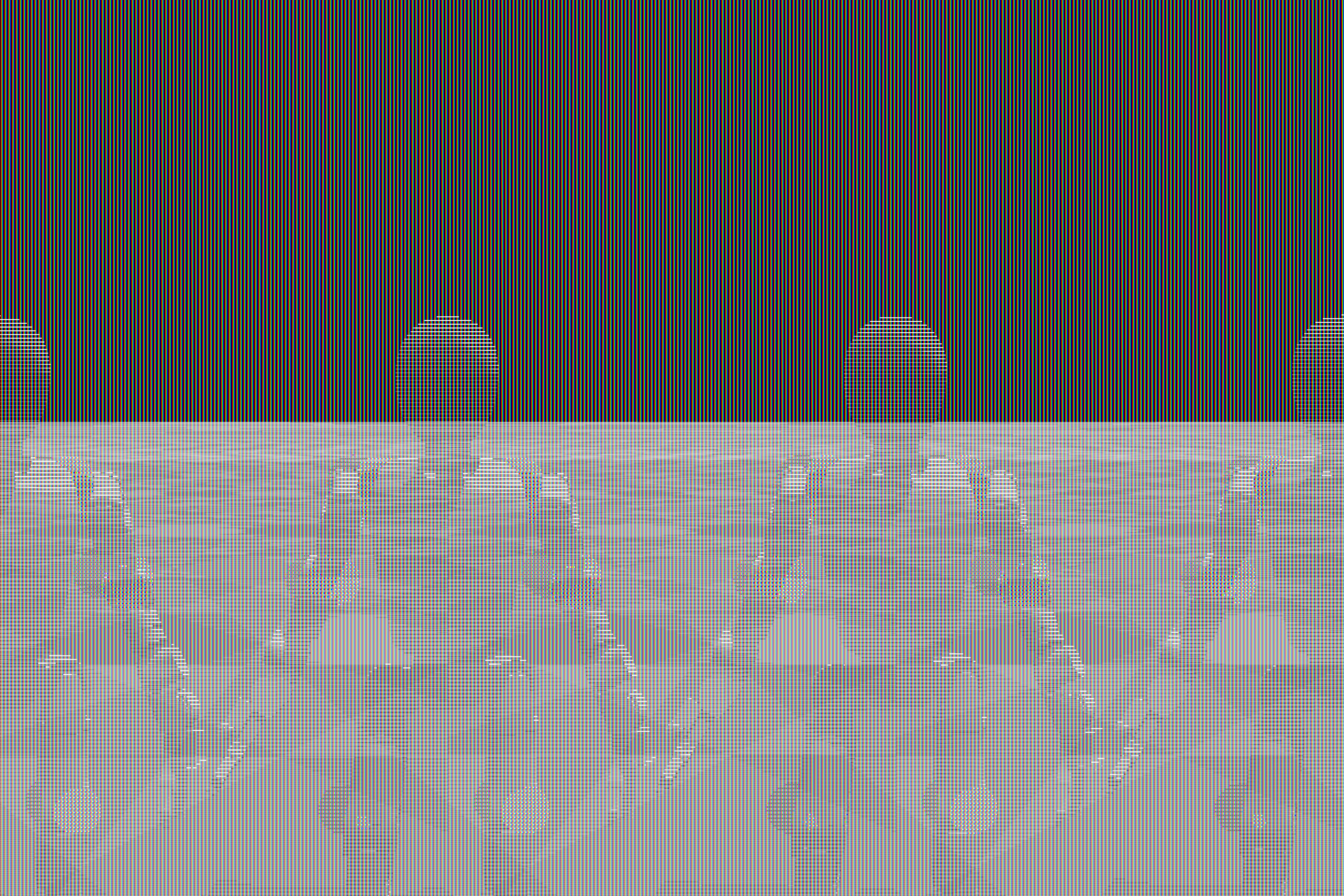}};
        \hspace{0pt}%
        \node at (1.5,-5.2) {\includegraphics[trim={0.6cm 2cm 2cm 2cm}, clip, scale=0.07]{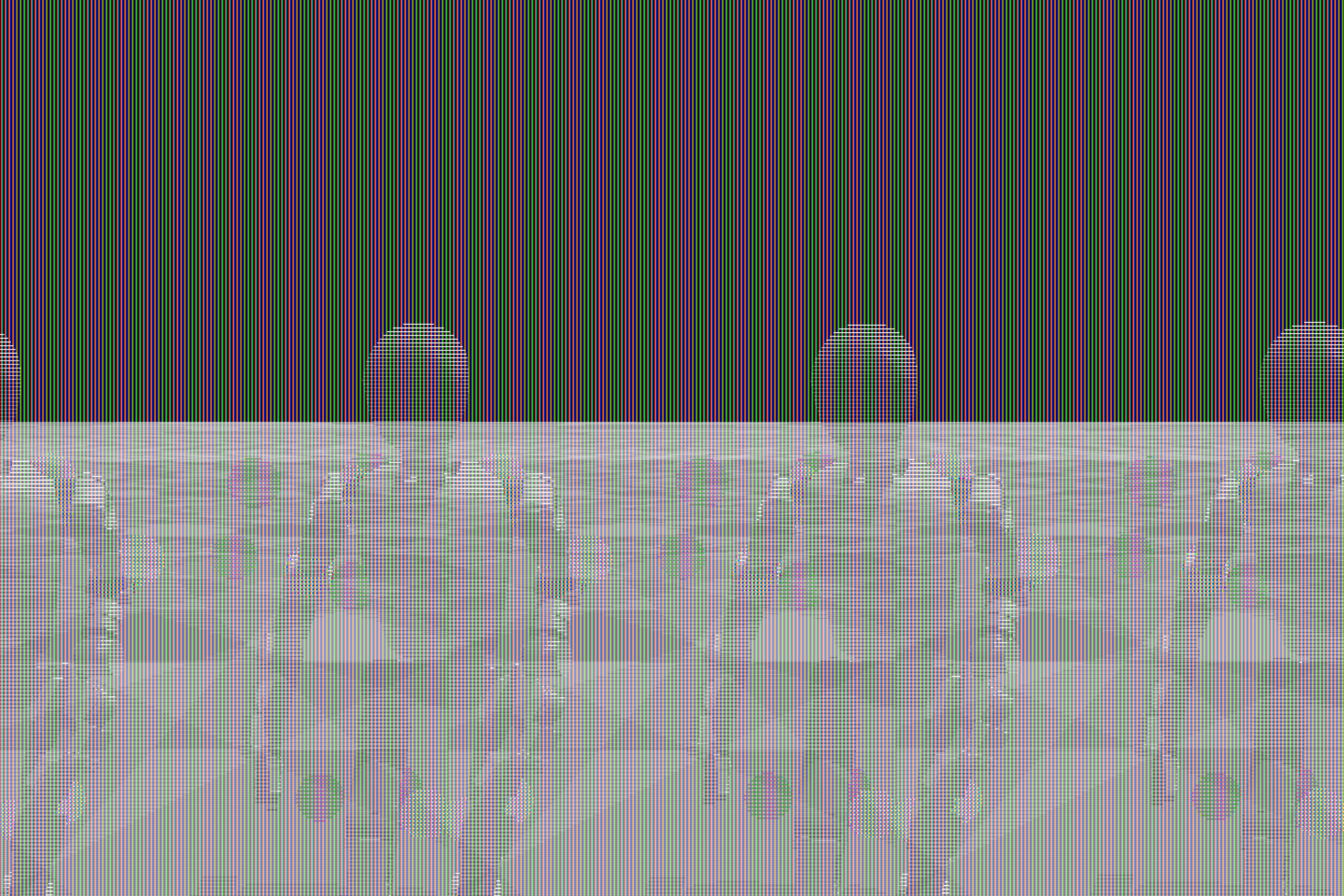}};
        \hspace{0pt}%
        \node at (5.0,-5.2) {\includegraphics[trim={0.6cm 2cm 2cm 2cm}, clip, scale=0.07]{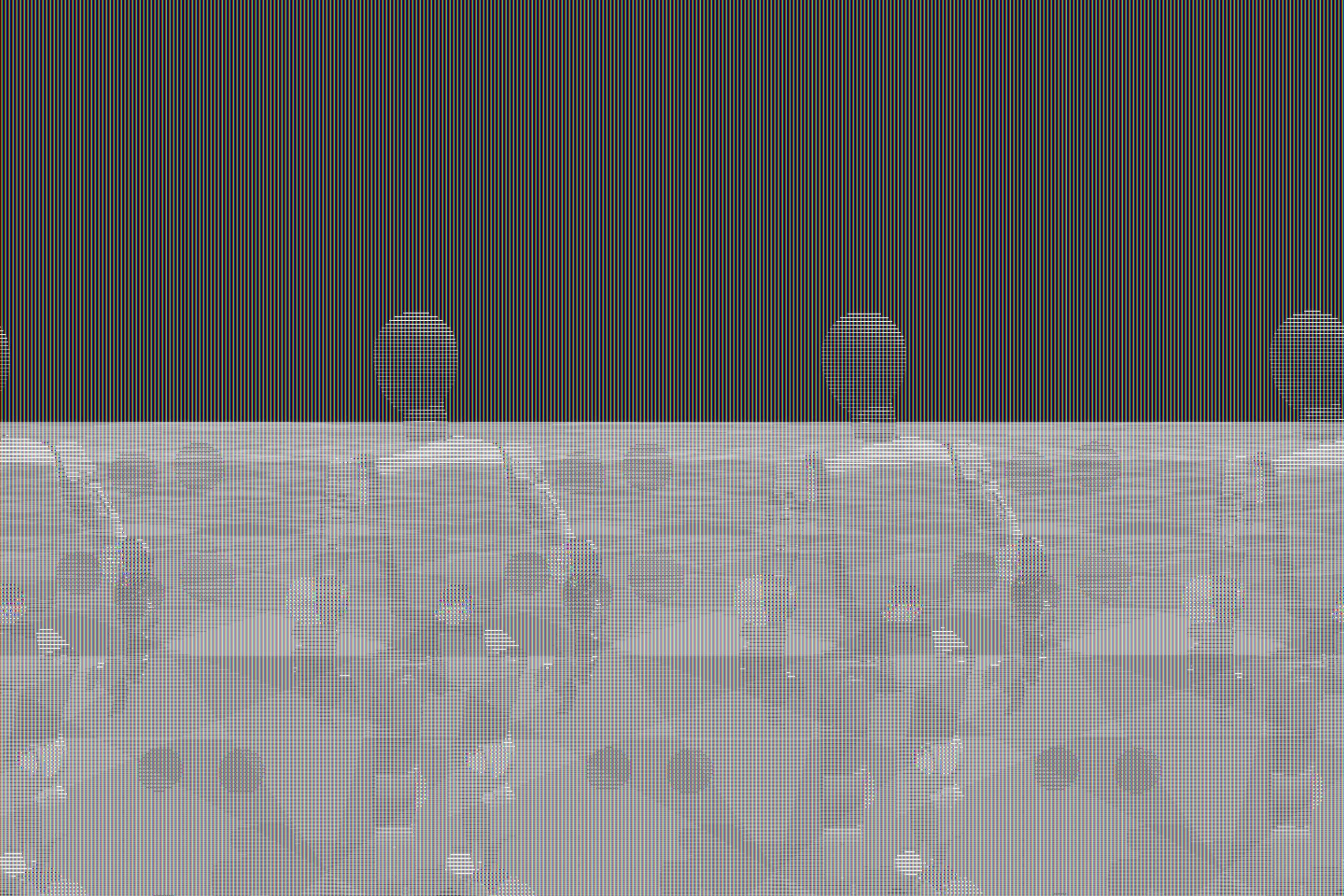}};
        \hspace{0pt}%
        \node at (8.5,-5.2) {\includegraphics[trim={0.6cm 2cm 2cm 2cm}, clip, scale=0.07]{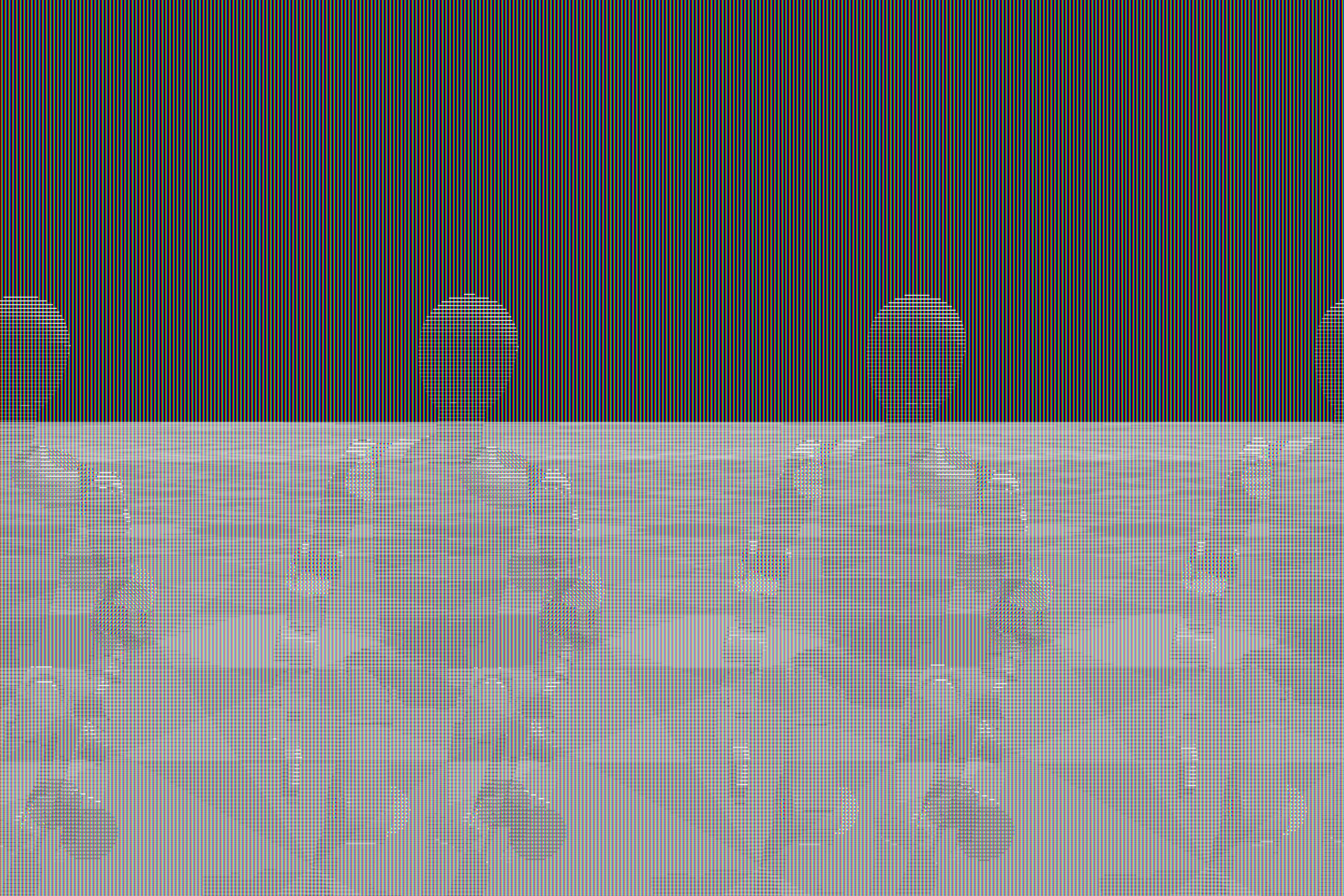}};
        \hspace{0pt}%
        \node at (12.0,-5.2) {\includegraphics[trim={0.6cm 2cm 2cm 2cm}, clip, scale=0.07]{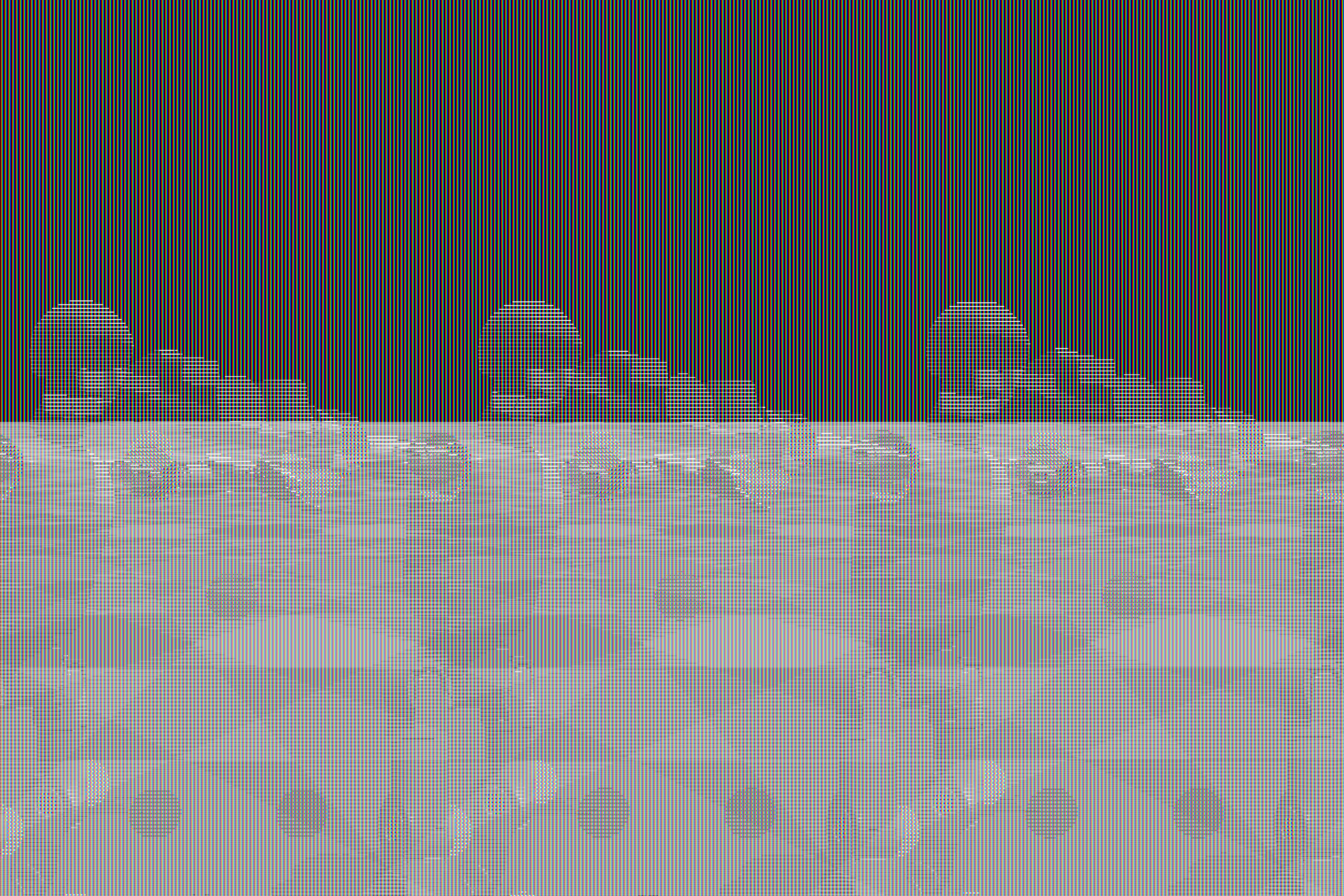}};
        
        \node[rotate=90, scale=0.6] at (-4.0,-7.4) {\fontfamily{lmss}\selectfont VENOM-clean-reduced};
        \node at (-2.0,-7.4) {\includegraphics[trim={0.6cm 2cm 2.6cm 2cm}, clip, scale=0.07]{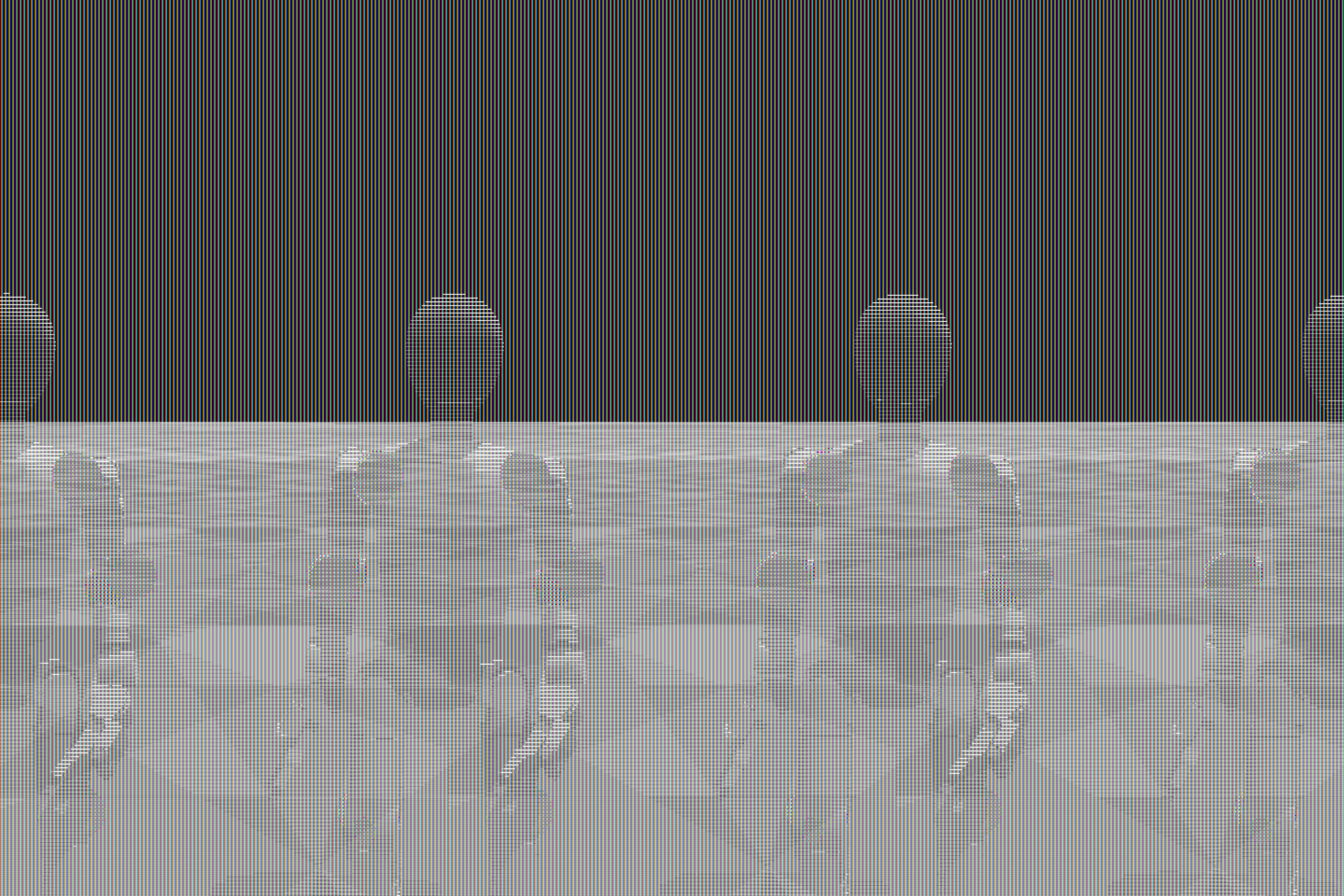}};
        \hspace{0pt}%
        \node at (1.5,-7.4) {\includegraphics[trim={0.6cm 2cm 2cm 2cm}, clip, scale=0.07]{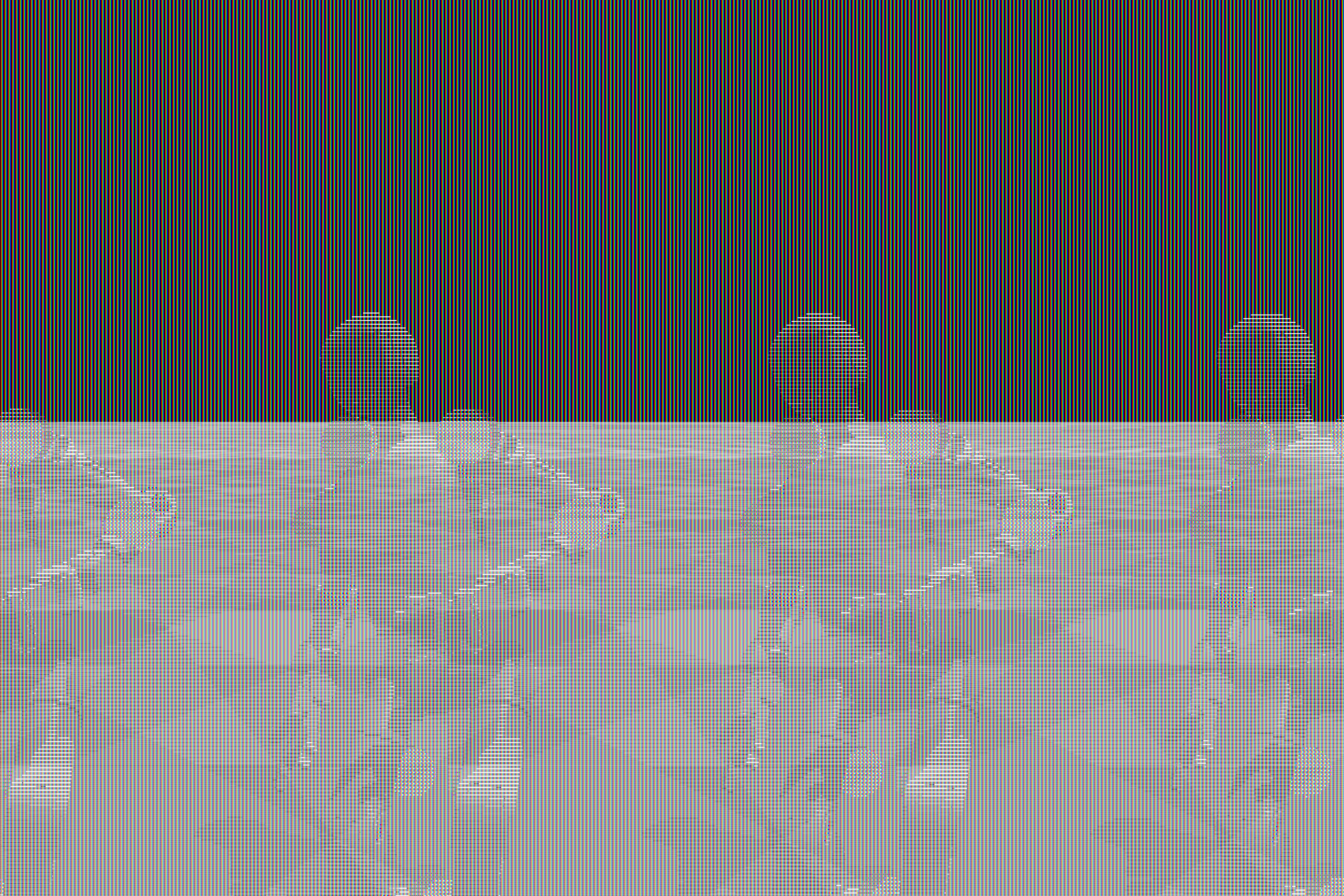}};
        \hspace{0pt}%
        \node at (5.0,-7.4) {\includegraphics[trim={0.6cm 2cm 2cm 2cm}, clip, scale=0.07]{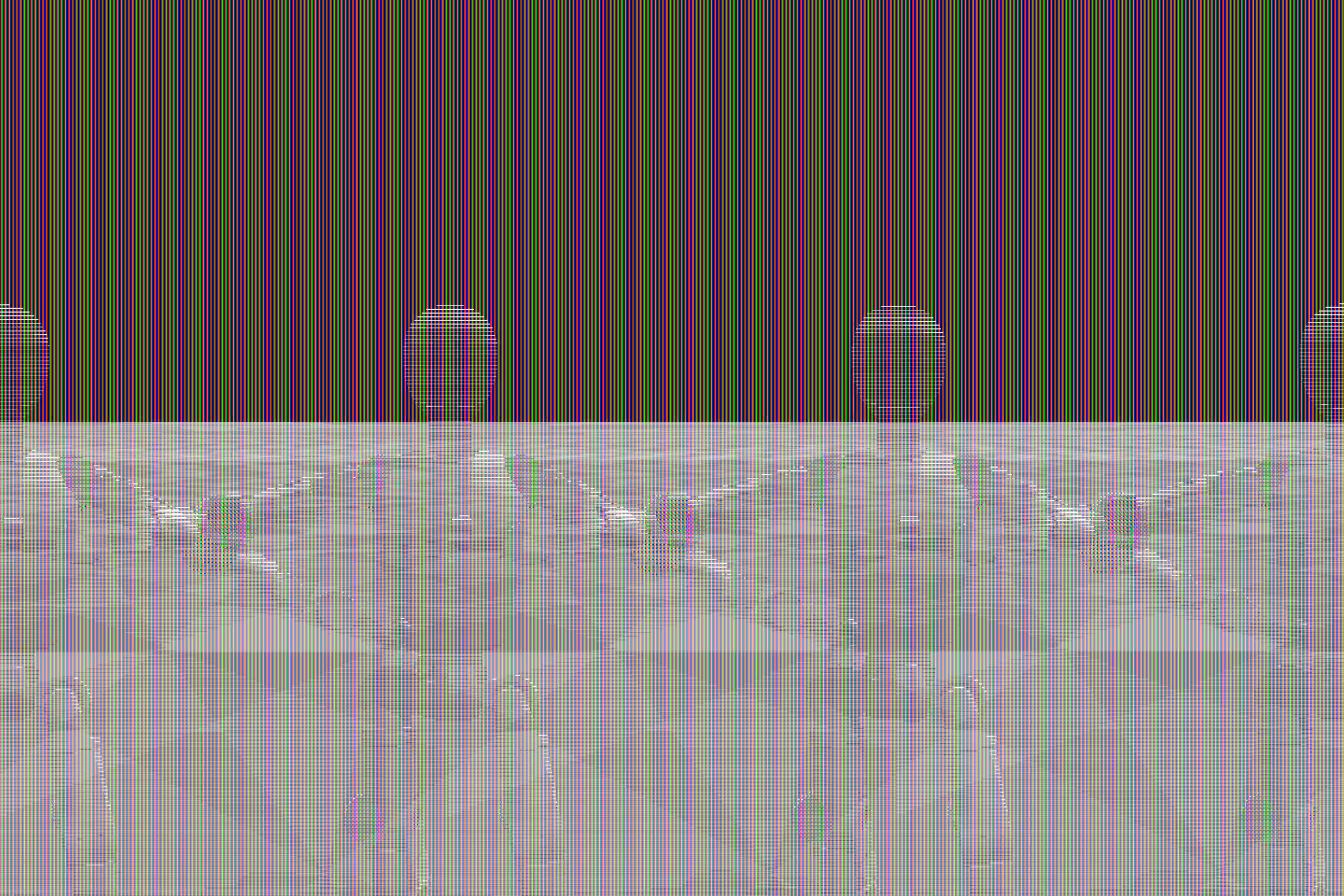}};
        \hspace{0pt}%
        \node at (8.5,-7.4) {\includegraphics[trim={0.6cm 2cm 2cm 2cm}, clip, scale=0.07]{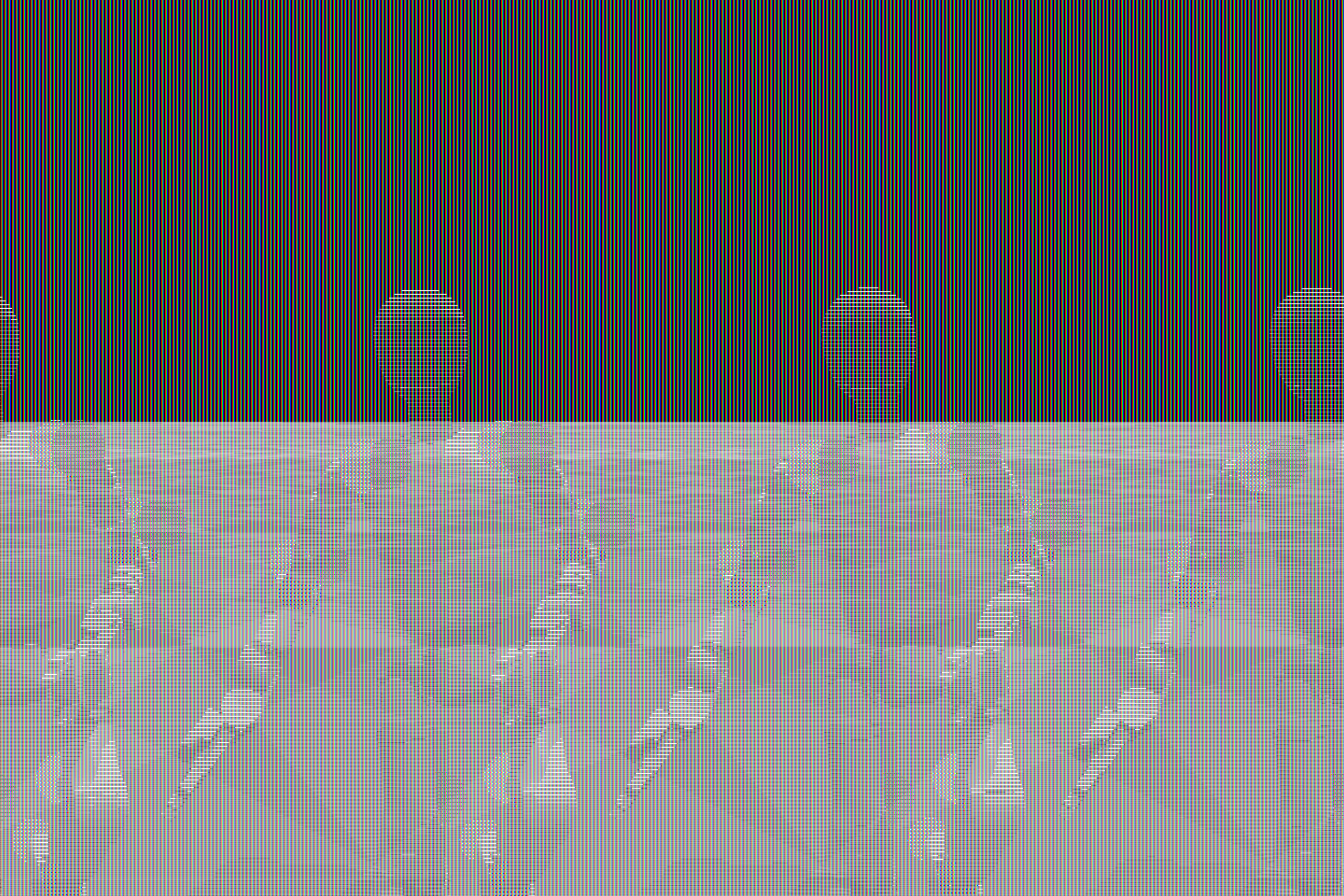}};
        \hspace{0pt}%
        \node at (12.0,-7.4) {\includegraphics[trim={0.6cm 2cm 2cm 2cm}, clip, scale=0.07]{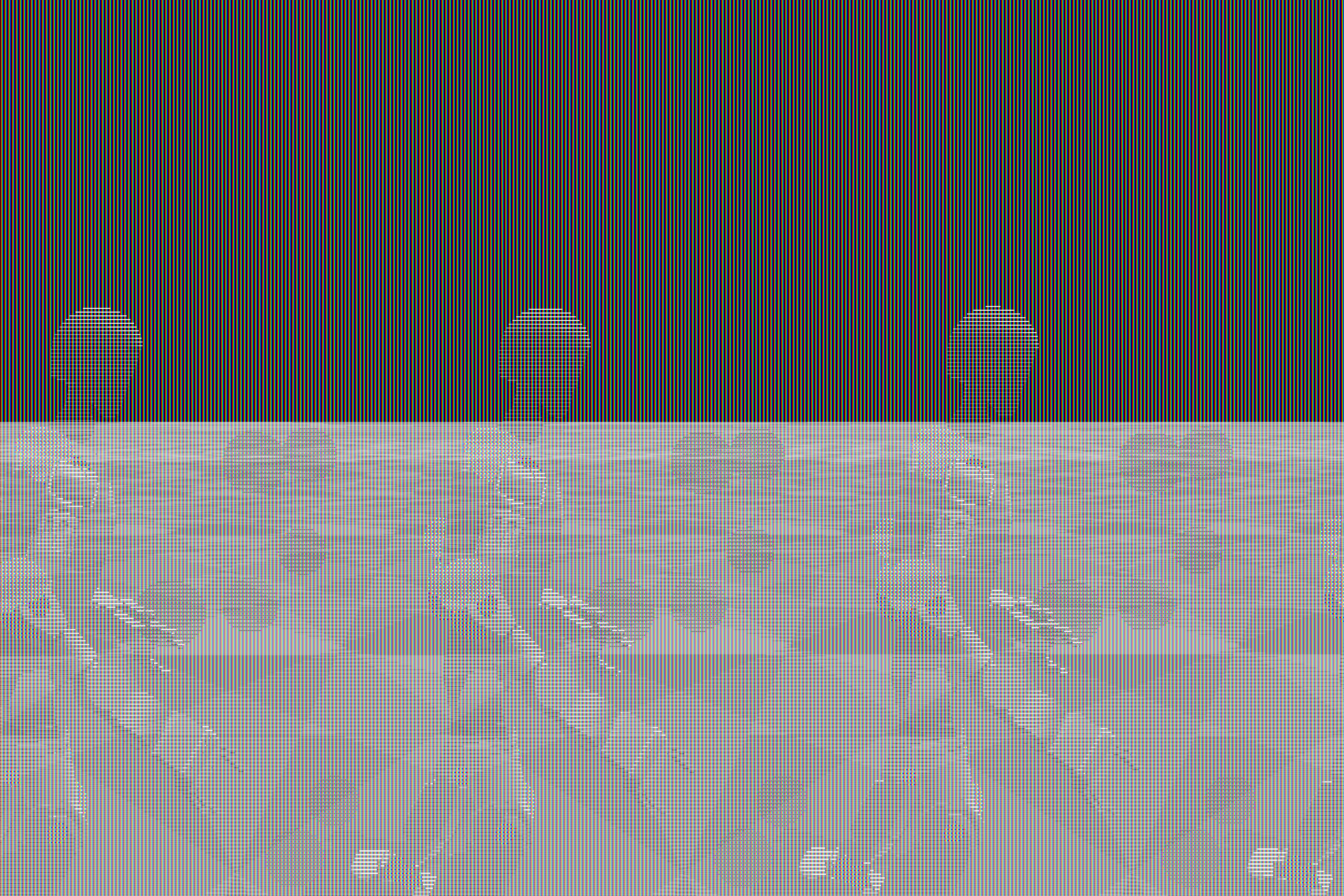}};
    \end{tikzpicture}

    \caption{The above frames are the rollouts of VENOM models for Booster T1 29 DoF for single-leg hopping and turn motion. From these frames, VENOM-noisy clearly tracks the motion comfortably compared the latter two, particularly VENOM-clean-reduced which eventually loses balance and falls.}
    \label{fig:boostert129dof}
\end{figure*}
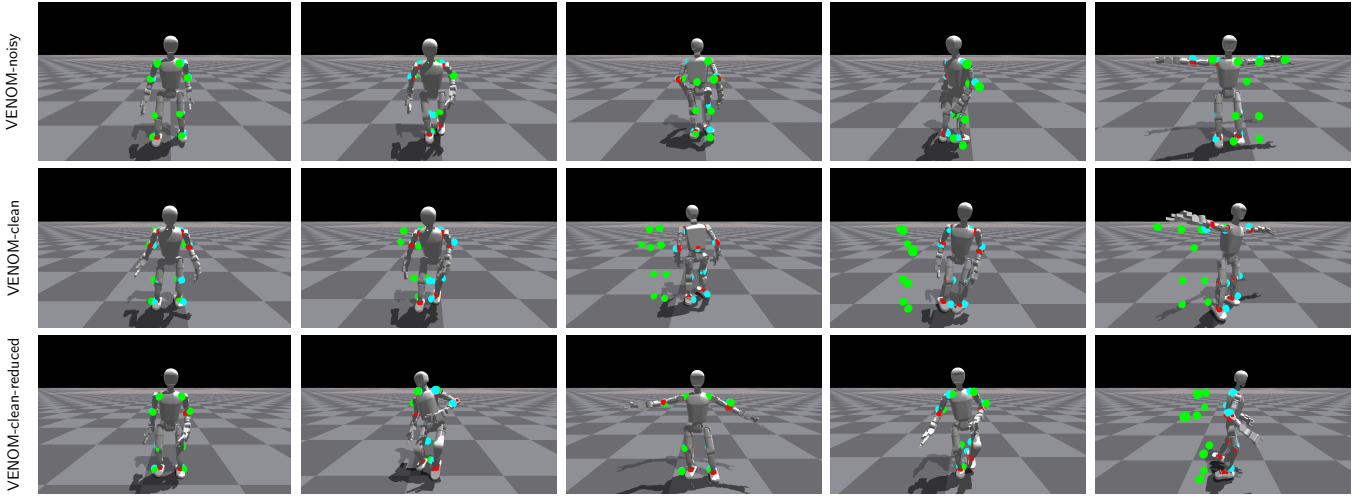

\section{Results and Discussion}
We trained VENOM on the VENOM dataset with and without noisy rollouts, which are named VENOM-clean and VENOM-noisy, respectively. We also trained VENOM on a quarter of the dataset without noise which is named VENOM-clean-reduced. The other baselines used for comparison are TWIST2 models trained on one robot from the VENOM dataset (Unitree G1 was chosen for this), and trained on the entire VENOM dataset without noise, which are named as TWIST2-clean-G1 and TWIST2-clean, respectively. The difference between TWIST2 models and the experts is that the TWIST2 models are trained in a supervised manner on the VENOM dataset with the same architecture implemented in TWIST2's framework whereas the experts are trained in TWIST2's RL training framework itself. Sample frames that show VENOM-noisy deployed on multiple humanoids in simulation are shown in Fig. \ref{fig:screenshots}.

\subsection{Scaling across bodies and noise}
We can observe improvement in performance in three directions; change in architecture, increasing number of robots in training data, and introducing noisy data both in visual demonstrations and metric evaluation.

We observe that using the same architecture used to train the TWIST2 motion tracking expert for G1, TWIST2-clean-G1, matches the expert's motion tracking capability. However, upon increasing the number of robots in the training data, TWIST2-clean displays lack in motion tracking capability and fails to stabilize the robots while tracking reference motion. This is due to negative transfer and the training was dominated by Unitree G1's data.

When we switch from MLP to GPT-based architecture, the VENOM models outperform TWIST2-clean. VENOM-clean tracks the reference motion better than VENOM-noisy-G1, indicating that increasing robot data increases motion tracking capability. Overall, VENOM-noisy displays the best motion tracking capability among all the trained models, indicating that including noisy rollouts help expose the policy to underexplored observation regions and in turn enables robustness.

\begin{table}
    \centering
    \caption{Metric Evaluation: Joint Tracking Error between Predictions and Targets in Unitree G1}
    \label{tab:metricevaluation}
    \begin{tabular}{c|c}
        \hline
        Model & Root Mean \\
        & Square Tracking \\
        & Error (radians) \\
        \hline
        VENOM-noisy & \textbf{0.1052} \\
        VENOM-clean & 0.1059 \\
        VENOM-clean-reduced & 0.1181 \\    
        VENOM-noisy-G1 & 0.1571 \\
        TWIST2-clean & 0.1686 \\
        TWIST2-clean-G1 & 0.1067 \\
        \hline
    \end{tabular}
\end{table}

\begin{table*}
    \centering
    \caption{Metric Evaluation: Joint Tracking Accuracy and Motion Tracking Stability across different Behavior Categories}
    \label{tab:jointrackingerror}
    \begin{tabular}{c|c|c|c|c|c|c}
        \hline
        Model & Robot & \multicolumn{2}{c|}{Joint Tracking Accuracy} & \multicolumn{2}{c|}{Motion Tracking Stability} & Success \\
        \cline{3-7}
        & & RMSE & P95 & Absolute Root Error & P95 & Average Rate \\
        & & (radians) & (radians) & Rate (cm/s) & (cm/s) & Percentange\\
        \hline
        VENOM-noisy & Unitree G1        & $0.4237_{\pm 0.0812}$ & $0.5539$ & $4.8025_{\pm 4.2905}$ & $12.3560$  & $96\%$ \\
                    & Unitree H1        & $0.5781_{\pm 0.1180}$ & $0.7791$ & $8.0430_{\pm 21.4745}$ & $17.8905$ & $92\%$ \\
                    & Unitree H1 v2     & $0.4518_{\pm 0.0636}$ & $\textbf{0.5286}$ & $5.9015_{\pm 4.9815}$ & $\textbf{13.2115}$ & $96\%$ \\
                    & Fourier N1        & $0.4820_{\pm 0.0820}$ & $\textbf{0.6306}$ & $3.6120_{\pm 3.1765}$ & $\textbf{8.8945}$ & $100\%$ \\
                    & Booster T1 29 DoF & $0.4585_{\pm 0.0717}$ & $\textbf{0.5753}$ & $6.9105_{\pm 3.9580}$ & $12.2250$ & $100\%$ \\
        \hline
        VENOM-clean & Unitree G1        & $0.3975_{\pm 0.0809}$ & $0.5586$ & $1.3962_{\pm 5.6715}$ & $14.0460$ & $100\%$ \\
                    & Unitree H1        & $0.5784_{\pm 0.1508}$ & $0.8616$ & $0.9962_{\pm 1.3071}$ & $\textbf{12.3025}$  & $100\%$ \\
                    & Unitree H1 v2     & $0.4582_{\pm 0.0690}$ & $0.5672$ & $3.6852_{\pm 19.1314}$ & $23.1770$  & $92\%$ \\
                    & Fourier N1        & $0.5087_{\pm 0.1458}$ & $0.7958$ & $1.2087_{\pm 1.4266}$ & $14.3035$  & $100\%$ \\
                    & Booster T1 29 DoF & $0.4704_{\pm 0.0855}$ & $0.6466$ & $1.9777_{\pm 1.2991}$ & $19.4050$  & $92\%$ \\
        \hline
        VENOM-clean & Unitree G1        & $0.4117_{\pm 0.0820}$ & $\textbf{0.5515}$ & $10.8785_{\pm 17.6850}$ & $22.6795$ & $100\%$ \\
        -reduced    & Unitree H1        & $0.5557_{\pm 0.1523}$ & $0.8188$ & $5.0900_{\pm 9.6370}$ & $12.3610$ & $96\%$ \\
                    & Unitree H1 v2     & $0.5144_{\pm 0.0878}$ & $0.6937$ & $8.2000_{\pm 12.0210}$ & $19.2130$ & $88\%$ \\
                    & Fourier N1        & $0.5126_{\pm 0.0946}$ & $0.6652$ & $4.9020_{\pm 7.3895}$ & $11.5415$  & $100\%$ \\
                    & Booster T1 29 DoF & $0.4547_{\pm 0.0756}$ & $0.6193$ & $11.4095_{\pm 35.1475}$ & $23.8295$  & $76\%$ \\
        \hline
        VENOM-noisy-G1 & Unitree G1         & $0.4608_{\pm 0.0918}$ & $0.6715$ & $11.2585_{\pm 5.5600}$ & $19.2280$  & $96\%$ \\
        \hline
        Expert & Unitree G1        & $0.4105_{\pm 0.0982}$ & $0.6028$ & $3.8490_{\pm 4.2330}$ & $\textbf{10.0030}$ & $100\%$ \\
               & Unitree H1        & $0.5749_{\pm 0.1100}$ & $\textbf{0.7583}$ & $10.0042_{\pm 42.1800}$ & $18.4390$  & $100\%$ \\
               & Unitree H1 v2     & $0.4653_{\pm 0.0855}$ & $0.6023$ & $9.8450_{\pm 7.0730}$ & $23.4015$  & $100\%$ \\
               & Fourier N1        & $0.4862_{\pm 0.1035}$ & $0.6546$ & $4.9570_{\pm 5.8645}$ & $11.6335$  & $100\%$ \\
               & Booster T1 29 DoF & $0.4482_{\pm 0.0845}$ & $0.5788$ & $3.3210_{\pm 6.6265}$ & $\textbf{9.1030}$  & $100\%$ \\
        \hline
    \end{tabular}
\end{table*}

\subsection{Motion Tracking}
All models were evaluated on their ability to track reference motions for one humanoid (Unitree G1 was chosen) to evaluate the multi-tasking capability. For the metric evaluation done in Table \ref{tab:metricevaluation}, each model was evaluated on 10 random motions sampled from the VENOM dataset for each model and the root mean squared error between the desired joint positions and joint observations were calculated. It can be observed that overall in terms of joint tracking the VENOM models perform better than TWIST2-clean. From this evaluation, all the VENOM models show comparable performances despite changes in training data such as introducing noise and changing size of the dataset. TWIST2-clean fails to track the reference motion as confidently as the other models due to negative transfer.

Unlike evaluating models that predict motion in a 3-D kinematic space rather than in a realistic physics environment like motionGPT\cite{motiongpt}, comparing ground truth with generated motion at joint level does not suffice to evaluate our models as it does not capture the tracking stability of the resulting humanoids' motions. Hence, we further extend the evaluation by evaluating joint tracking accuracy and motion tracking stability in 3-D Cartesian space for different behavior categories that contain both acyclic and cyclic movements: athletic, crouching, dancing, hopping, and locomotion as tabulated in Table \ref{tab:jointrackingerror} (five random motions were sampled from the dataset in each behavior category).

Similar to Table \ref{tab:metricevaluation}, joint tracking accuracy was calculated by using the root mean squared error between the target and generated joint positions. For the motion tracking stability, first the L2 norm is calculated between the reference root position and the observed root position of the humanoid. Then the rate of absolute changes in the L2 norm is calculated and tabulated in Table \ref{tab:jointrackingerror}. The root position error, $e_t$  and absolute root error rate, $r_t$, are defined as
\begin{align}
    e_t &= \left\|\mathbf{p}^{\mathrm{ref}}_t - \mathbf{p}^{\mathrm{gen}}_t \right\|_2, \\
    r_t &= \left|e_t - e_{t-1} \right|.
\end{align} The reason why the absolute root error rate was chosen instead of error itself is because there is inevitable drift between the reference root and actual root of the robot due to resulting learned dynamics and the robot morphology (also the motion tracking task is driven by the joint targets not by root targets). Consequently, the change in error rate provides a meaningful measure of whether the generated root follows similar movement pattern as the reference motion despite any drift. The success rate is the percentage of motions completed by the humanoids out of total number of evaluated motions, which is 25 (for each model it is 5 motions for each behavior category and robot). While the success rates of the models are all above $90\%$, events such as losing stability, excess drift, occasional poor joint tracking, prolonged stationary behaviors, produce large deviations from the reference root. To avoid these infrequent events to dominate the evaluation, we report the 95th percentile as the deciding metric.

These tables show that although VENOM-noisy tracks motion slightly better than the other models, and overall all the models show similar capabilities in joint tracking, experts that were trained using reinforcement learning achieve higher motion stability as shown from the success rates, signaling that perhaps there is a compromise between motion tracking and motion stability when implementing either supervised learning or reinforcement learning. From the standard deviation values in the motion tracking stability column in Table \ref{tab:jointrackingerror}, VENOM-noisy shows the least change in drift from reference root signaling at compromising motion stability with overall motion tracking. However, upon visually inspecting the simulations, all models struggle at times to track highly dynamic motions or motions that involve feet losing contact with the ground such as jumping. Enabling the models to cover these behaviors much better might require a different training curriculum or approach. As an example, Fig. \ref{fig:boostert129dof} shows that VENOM-noisy comfortably generates actions for Booster T1 29 DoF for a single-leg hopping and turning motion that targets the reference motion well compared to VENOM-clean and VENOM-clean-reduced.

It is worth noting, that although the experts were trained in a asymmetric actor-critic setting with access to privileged information during training, VENOM achieves comparable performance despite having access to limited observation available to the policy. This shows that distillation via supervised learning can recover much of the expert's capability with a compromise in tracking stability.

\section{Discussion}
\subsection{Limitations}
Ultimately, VENOM-noisy despite being trained on multiple noisy robot data meets similar capabilities but does not surpass the RL experts' performances to a significant extent. Highly dynamic motions such as hopping, sprinting, jumping, are hard to track. All the models, including the RL experts, eventually learn behaviors that prevent losing contact with the ground to ensure stability.

\subsection{Future Work}
The VENOM-dataset was curated as a first step under current constraints to see if we can develop a model for multi-humanoid full-body tracking. Dynamic motions that involve flight phases are still difficult to imitate due to lack of a prediction horizon. Future work will investigate prediction models that can better capture long-term dependencies of these kind of motions. To build a much more effective model, it would be prudent to directly train humanoids in simulation using reinforcement learning frameworks for motion tracking and imitation like TWIST2 \cite{twist2} instead of supervised learning. Scaling across arbitrary humanoid embodiments rather than just the existing ones, as demonstrated in XHugWBC\cite{xhugwbc} and GENBOT-1K \cite{towards}, would result in a model much more capable of generalizing to unseen humanoids and motions. To facilitate zero-shot transfer to unseen humanoids, it would be better to make VENOM robot-agnostic by using algorithms such as URMA embedding\cite{bohlinger2024policyrunallendtoend}. One significant issue is to retarget motion data to each humanoid prior to training which can get quite cumbersome. By using online retargeting policies such as ReActor \cite{reactor} and extending it to generalize across different morphologies, it might be possible to circumvent this issue.

\subsection{Miscellaneous}
VENOM was also trained to autoregressively predict motion and we were able to observe the impact of conditioning inputs with joint targets on the policy's ability to control. VENOM was trained for motion generation by removing all input related to tracking, leaving out just the proprioception data as input to the model (VENOM-motion-gen). VENOM was also trained by randomly masking the motion tracking input during training except for the proprioception data (VENOM-motion-gen-mask). Despite using the same training approach to autoregressively predict motion used in MoCapAct\cite{mocapact}, HMG\cite{padmanabhan}, and LocoGPT\cite{locogpt}, both VENOM-motion-gen and VENOM-motion-gen-mask fail to stabilize robots. VENOM-motion-gen-mask stabilizes the humanoids by masking the joint target input, but fails to track once the joint targets are unmasked. We hypothesize that motion generation and motion control may not scale identically with embodiment diversity. As the diversity of robots and behaviors increases, conditioning information becomes increasingly important for maintaining stability and preventing excess change in drift.

\section{Conclusion}
We propose a whole-body GPT-based multi-humanoid motion tracker called VENOM. VENOM unlike other SOTA cross-embodiment models, does not decouple upper and lower body control and can demonstrate a wider variety of expressive motions. We further contribute by curating a large low-level multi-embodiment motion tracking dataset called VENOM dataset using IsaacGym\cite{isaacgym} upon which we train VENOM. From the evaluation, VENOM-noisy consistently outperforms the baselines in terms of motion joint and root tracking, and root stability, though it does not exceed the combination of tracking and stability of RL experts. Although some motions are dynamically tough to track, VENOM learns to imitate without compromising stability despite lack of reward feedback. VENOM would be much more appreciated if trained on a wider range of embodiments, made to be robot-agnostic, and trained in a reinforcement learning setting.

\bibliographystyle{IEEEtran}
\bibliography{main}

\end{document}